\documentclass{article}

\usepackage{microtype}
\usepackage{graphicx}
\usepackage{subcaption}
\usepackage{booktabs} 

\usepackage{hyperref}

\newcommand{\etal}{\textit{et~al.}}

\newcommand{\bigdot}[1][0.4ex]{%
  \tikz[baseline=-0.6ex]\fill (0,0) circle (#1);%
}

\usepackage[accepted]{icml2026}

\usepackage{amsmath}
\usepackage{amssymb}
\usepackage{mathtools}
\usepackage{amsthm}

\usepackage{alltt}
\usepackage{enumitem}
\usepackage{tcolorbox}
\usepackage{multirow, xcolor, array, tikz} 
\usepackage[dvipsnames]{xcolor}
\definecolor{cvprblue}{rgb}{0,0.08,0.45}

\usepackage[capitalize,noabbrev]{cleveref}

\theoremstyle{plain}
\newtheorem{theorem}{Theorem}[section]
\newtheorem{proposition}[theorem]{Proposition}
\newtheorem{lemma}[theorem]{Lemma}
\newtheorem{corollary}[theorem]{Corollary}
\theoremstyle{definition}

\newtheorem{assumption}[theorem]{Assumption}
\theoremstyle{remark}

\usepackage[textsize=tiny]{todonotes}

\icmltitlerunning{Neutral-Reference Prompting for Vision–Language Models}

\begin{document}

\twocolumn[
  \icmltitle{Neutral-Reference Prompting for Vision–Language Models}



  \icmlsetsymbol{equal}{*}

  \begin{icmlauthorlist}
   \icmlauthor{Senmao Tian}{yyy}
    \icmlauthor{Xiang Wei}{yyy}
    \icmlauthor{Shunli Zhang}{yyy}
  \end{icmlauthorlist}

  \icmlaffiliation{yyy}{Beijing Jiaotong University, Beijing, China}

  \icmlcorrespondingauthor{Shunli Zhang}{smtian1204@gmail.com}

  \icmlkeywords{Machine Learning, ICML}

  \vskip 0.3in
]



\printAffiliationsAndNotice{}  

\begin{abstract}
Efficient transfer learning of vision–language models (VLMs) commonly suffers from a Base–New Trade-off (BNT): improving performance on unseen (new) classes often degrades accuracy on known (base) classes.   Addressing how to boost recognition of unseen classes without sacrificing known-class performance remains a central challenge. Existing work often simplistically attributes the BNT to overfitting on known classes.
We observe an interesting phenomenon: VLMs frequently exhibit asymmetric confusion on certain downstream data, i.e., samples of class A are systematically mispredicted as class B, while the reverse confusion (B → A) rarely occurs.  For known classes, this kind of bias can be mitigated by tuning using a cross-entropy loss, but for unseen classes, such pretraining-induced bias persists and harms generalization.
Motivated by this, we propose NeRP, a plug-and-play prompting correction strategy that improves discrimination on unseen classes without modifying model parameters. NeRP leverages neutral text prompts and reference images to measure class-wise prior preferences along the pre-trained inter-class geometry, and combines them with the sample likelihood to obtain the model’s surrogate score. If, for a given sample, the prior strongly favors the current prediction while the observed evidence is clearly insufficient, we perform a local flip between easily confusable class pairs, thereby correcting prior-dominated mispredictions.
Extensive experiments across multiple backbones and 15 few-shot and cross-domain benchmarks show that NeRP substantially improves accuracy on unseen classes while preserving known-class prediction performance. Code is available at \url{https://github.com/Sheldon04/NeRP}.
\end{abstract}

\begin{figure}[t]
  \centering
   \includegraphics[width=0.95\linewidth]{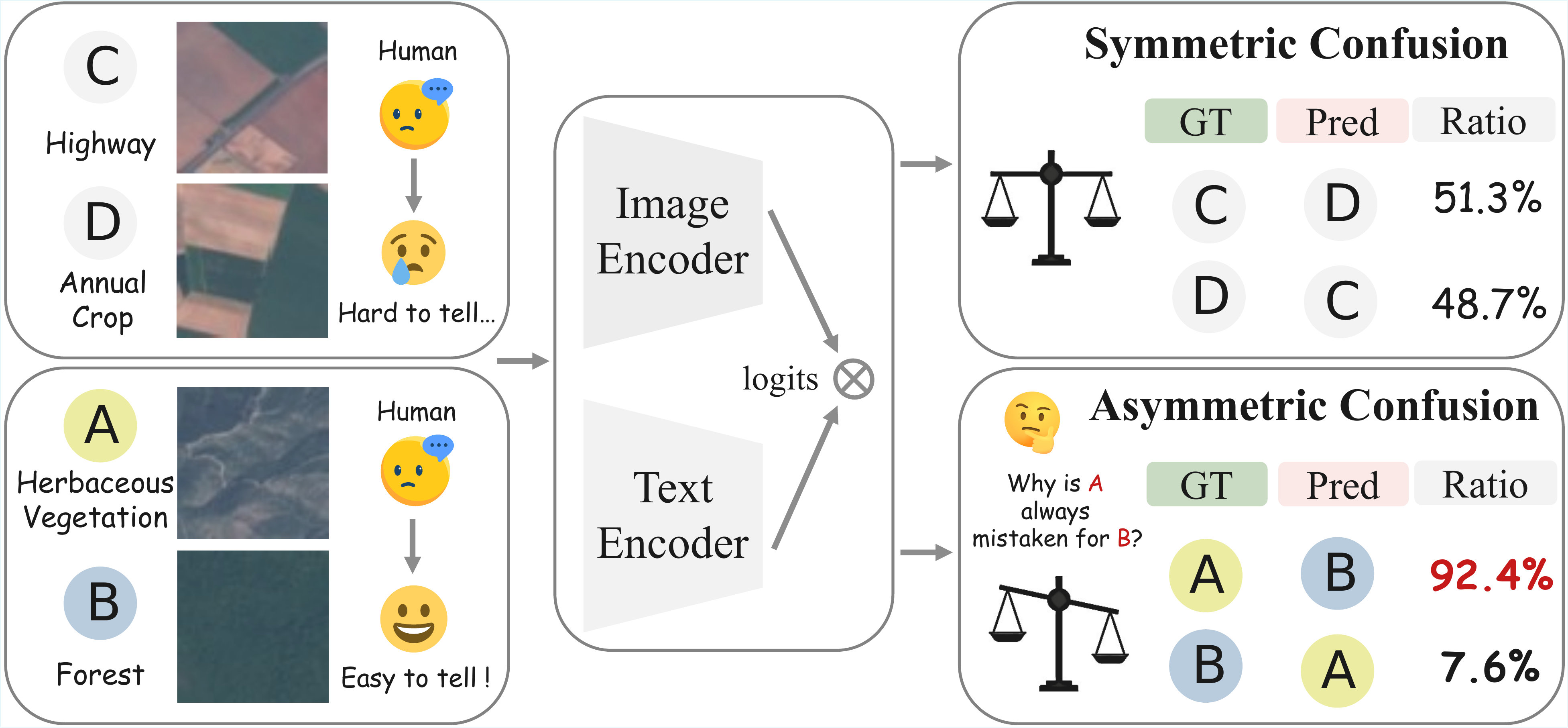}
   \caption{We observe two distinct types of confusion on downstream data. The first is symmetric confusion: for two similar classes the model struggles to discriminate them, samples from each class are misclassified as the other at comparable rates. The second is asymmetric confusion: the two classes may not be intrinsically ambiguous, yet the model systematically mislabels class A as B far more often than it mislabels B as A. More examples can also be found in \cref{fig:abconsufion}.}
   \label{fig:motivation}
   \vspace{-0.4cm}
\end{figure}

\section{Introduction}
\label{sec:intro}
Pre-trained VLMs \cite{clip, align, flamingo, filip, vila,a2,a5,a6} have demonstrated remarkable zero-shot capabilities in few-shot and cross-domain recognition tasks. However, prompt tuning and lightweight adaptation methods often suffer from the BNT problem \cite{dept,cocoop,a1,a3,a4}: improving performance on unseen (new) classes typically comes at the cost of degraded accuracy on seen (base) classes. Previous studies commonly attribute this phenomenon to overfitting \cite{dpc, dept, promptsrc, promptkd, kgcoop, surpl, nlprompt} on base classes during fine-tuning, which compromises generalization to unseen classes. To mitigate overfitting, various methods have been explored, such as modifying loss functions \cite{promptsrc, tcp}, constraining prompt learning \cite{tcp, cocoop, coprompt, proda}, enhancing multimodal interaction \cite{mma, maple, mmrl}, and incorporating external knowledge \cite{promptkd, caspl}. 


However, our investigation reveals that poor performance on unseen classes cannot be fully explained by overfitting alone. A more subtle issue is the asymmetric confusion as \cref{fig:motivation} shows. This arises from data imbalance during VLM pre-training \cite{intrinsic, sober, closer}; therefore, on both the image side and the text side, there exists a certain preference for some classes. During fine-tuning, the cross-entropy loss helps the model learn correct class boundaries \cite{tian2026sampling}, effectively suppressing biases for seen classes. Nevertheless, for unseen classes, existing approaches \cite{coop, clip-adapter, prograd} heavily rely on the inherent zero-shot ability of vanilla VLMs. Consequently, the asymmetric biases inherited from pre-training become more pronounced on unseen classes, severely limiting generalization.

Identifying confusable class pairs and reversing the model's bias is appealing, but two issues remain: the preferred direction of confusion is unknown, and naïve flipping can harm correct predictions unless biased errors are detected with high confidence.

To address the above challenges, we first revisit how VLMs are efficiently adapted to downstream image classification \cite{clip, transfer2, transfer3}. Whether via prompt-based tuning or lightweight adapters, existing approaches essentially use learnable parameters to fit the statistics of the downstream data. These methods either fine-tune the text encoder or feed fine-tuned prompt vectors into a vanilla text encoder to produce a text embedding for each class; at inference, the image encoder embeds an unseen image (optionally with a learnable image prompt) and computes cosine similarities against all class embeddings to obtain the classification logits. For any image containing semantic content, the model naturally selects the text semantics closest to the image semantics as the predicted class. However, what should the model output for an image devoid of semantic information? This motivates our concept of \emph{Neutral-Reference Prompt} (NeRP).

We define a \emph{neutral-reference prompt} as one that can describe any sample in the dataset while remaining class-agnostic: a neutral-reference image preserves only domain-specific style with semantic content removed, and a neutral text prompt contains no class-specific cues. Our theoretical and empirical analyses show that fine-tuning mainly deforms a low-dimensional subspace spanned by prototypes of known classes, while the zero-shot geometry among unseen classes remains largely stable. Consequently, for any pair of unseen classes, the neutral prior gap induced by the neutral-reference prompt and the model’s expected logit margin have the same sign with high probability. For an unbiased model, this gap is close to zero; otherwise, neutral-prior probing reveals the direction and strength of bias between confusable classes. We then construct Bayes-style posterior log-odds by adding this prior term to the observed sample logit margin. When the prior is strong but the evidence weak, we treat the prediction as prior-dominated and likely erroneous, and perform a local decision flip within the neighborhood of easily confusable classes; when both prior and evidence are weak, we regard the model as genuinely undecided and leave the prediction unchanged. Extensive experiments show that NeRP integrates seamlessly with diverse state-of-the-art prompt tuning and lightweight adaptation methods, consistently improving unseen-class performance on nearly all downstream benchmarks without retraining or modifying model parameters, thereby serving as a plug-and-play prompting strategy that balances discrimination and generalization in VLMs.

\section{Related Work}
\label{sec:related}

\subsection{Efficient Transfer Learning for VLMs.}
Vision-language models (VLMs) are designed to jointly understand images and text, with representative models like CLIP \cite{clip}, ALIGN \cite{align}, and Flamingo \cite{flamingo} showcasing powerful cross-modal capabilities. To further enhance these models' performance on specific tasks or domains, various transfer learning methods have been proposed to adapt their pre-trained representations. Early work, such as CoOp \cite{coop} replaces hand-crafted templates with learnable context vectors while freezing CLIP. CoCoOp \cite{cocoop} further conditions prompts on each image to reduce base-to-new drift. 
ProGrad \cite{prograd} updates prompts only when gradients align with general knowledge, and 
DPC  \cite{dpc} decouples optimization directions explicitly with a dual-prompt scheme. TCP \cite{tcp} enhances discriminative capability by incorporating semantic priors through mapping class-level textual knowledge into class-specific prompt tokens. In a similar vein, PromptSRC \cite{promptsrc} leverages tricks such as text diversification and self-ensembling to reduce overfitting, without incurring any inference-time cost. SkipT \cite{skipt} makes VLMs self-adapters—skip low-impact layers and classes to fine-tune faster with no extra params.

To better align multimodal representations, MaPLe \cite{maple} couples deep prompts in both branches via a learned projection. Visual-side prompting has also been strengthened with progressive designs such as ProVP \cite{provp} to stabilize feature distributions. MMA \cite{mma} inserts multi-modal adapters and shows that higher layers carry task-specific signals while lower layers preserve generalization.  MMRL \cite{mmrl} learns a shared representation space with extra tokens inserted only in higher layers, while regularizing the original class token to retain zero-shot priors, and uses a decoupled inference strategy.

\begin{figure*}[t]
  \centering
   \includegraphics[width=0.9\linewidth]{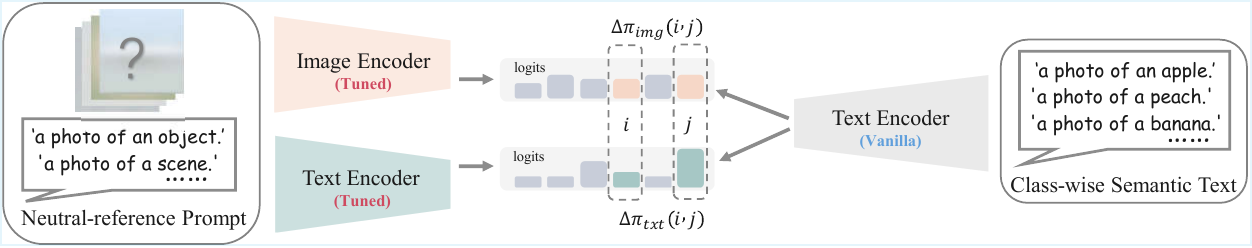}
   \caption{The pipeline of obtaining prior gaps from the neutral-reference prompt.}
   \label{fig:method}
   \vspace{-0.4cm}
\end{figure*}

\subsection{Bias in Pretrained VLMs}

Recent work has already studied biases in pretrained VLMs.  Hamidieh \etal~\cite{sobit} proposed the So-B-IT vocabulary and showed that CLIP over-associates harmful terms with certain demographics in open-vocabulary retrieval. 
Alabdulmohsin \etal~\cite{clipthebias} introduced Multi-Modal Moment Matching (M4) to balance data at both representation and association levels. Pushing further, Sahili \etal~\cite{sizenoteq} controlled model size, data scale, and data composition, and found that the “bigger and fairer” intuition does not generally hold.  Ghate \etal~\cite{intrinsic} quantified this more precisely: the choice of pretraining dataset is the dominant upstream factor.
Beyond social bias, broader forms of bias have been explored. Wang \etal~\cite{sober} built the CounterAnimal benchmark and revealed CLIP’s reliance on spurious cues such as background. Tu \etal~\cite{closer} revealed that CLIP’s shape bias weakens after ImageNet-targeted fine-tuning while texture reliance increases.
These insights also inform mitigation for downstream tasks. Li \etal~\cite{bgprompt} alleviate the overfitting caused by a single background class via learned background prompts and online mining. RS-CLIP \cite{rsclip} uses structured adaptation and data cleaning to ease class-imbalance effects, transferring generic CLIP to RS scenes. Motivated by these findings, we develop a general, efficient approach for VLMs that suppresses bias in transfer learning.



\section{Method}

\subsection{Preliminary}

\paragraph{Notation.}
Let $\mathcal{X}$ be the image space and $\mathcal{C}$ the label set with $|\mathcal{C}|=C$.
We write $d$ for the embedding dimension and assume L2-normalized features unless stated.
We explicitly distinguish raw encoders from normalized features:
let $g_{\mathrm{img}}:\mathcal{X}\!\to\!\mathbb{R}^d$ and $g_{\mathrm{txt}}:\text{Token}^*\!\to\!\mathbb{R}^d$ denote \emph{raw} image and text encoders, and define the normalized outputs
\[
f_{\mathrm{img}}(x):=\operatorname{norm}\!\big(g_{\mathrm{img}}(x)\big),
f_{\mathrm{txt}}(z):=\operatorname{norm}\!\big(g_{\mathrm{txt}}(z)\big).
\]
Let $g_{\mathrm{img}}^{0}, g_{\mathrm{txt}}^{0}$ be the corresponding \emph{pretrained (zero-shot)} raw encoders, and $f_{\mathrm{img}}^{0}:=\operatorname{norm}\!\circ g_{\mathrm{img}}^{0}$, $f_{\mathrm{txt}}^{0}:=\operatorname{norm}\!\circ g_{\mathrm{txt}}^{0}$ their normalized counterparts (CLIP-like).
For each class $c\!\in\!\mathcal{C}$ we define the class text prototypes
\[
\begin{aligned}
t(c) &:= \operatorname{norm}\big(g_{\mathrm{txt}}(\mathrm{prompt}(c))\big),\\
t^{0}(c) &:= \operatorname{norm}\big(g_{\mathrm{txt}}^{0}(\mathrm{prompt}_0(c))\big).
\end{aligned}
\]
Given an image $x\in\mathcal{X}$, the class score is
\(
\ell_c(x):=\,\langle f_{\mathrm{img}}(x),\, t(c)\rangle.
\)
We split the label set into base $\mathcal{B}$ (seen during fine-tuning) and novel $\mathcal{N}=\mathcal{C}\setminus\mathcal{B}$ (unseen).
Throughout, we fix the \emph{base subspace}
\[
S\;:=\;\mathrm{span}\{\,t^{0}(c):\ c\in\mathcal{B}\,\}\subset\mathbb{R}^d.
\]
For a dataset/domain $D$, let $\tau(D)$ denote the dataset-specific \emph{neutral-reference text prompt} (e.g., “a photo of an object.”).

\paragraph{Neutral priors.}
For the text side, we construct a zero-shot neutral vector
\[
u_{\mathrm{txt}}^{0}(D)\;:=\;\operatorname{norm}\!\big(g_{\mathrm{txt}}^{0}(\tau(D))\big).
\]
The per-class \emph{text prior logit} is
\begin{equation}
\label{eq:text-prior}
\pi_{\mathrm{txt}}(c;D)\;:=\;\big\langle t(c),\,u_{\mathrm{txt}}^{0}(D)\big\rangle, c\in\mathcal{C}.
\end{equation}

For the image side, we use the feature of the mean image (pixel-wise average of the preprocessed training images) as the \emph{neutral-reference image prompt} and then normalize:
\[
\begin{aligned}
\bar x^{D}\;&:=\;\frac{1}{n}\sum_{x\in \mathrm{train}(D)} \mathrm{preproc}(x),\\
u_{\mathrm{img}}(D)\;&:=\;f_{\mathrm{img}}(\bar x^{D})=\operatorname{norm}\!\big(g_{\mathrm{img}}(\bar x^{D})\big).
\end{aligned}
\]
Correlating with \emph{zero-shot} class prototypes yields the per-class \emph{image prior logit}
\begin{equation}
\label{eq:image-prior}
\pi_{\mathrm{img}}(c;D)\;:=\;\big\langle u_{\mathrm{img}}(D),\, t^{0}(c)\big\rangle, c\in\mathcal{C}.
\end{equation}
For a class pair $(i,j)$ we define the \emph{prior gaps}
\[
\begin{aligned}
\Delta\pi_{\mathrm{txt}}(i,j)\;&:=\;\pi_{\mathrm{txt}}(i;D)-\pi_{\mathrm{txt}}(j;D),\\
\Delta\pi_{\mathrm{img}}(i,j)\;&:=\;\pi_{\mathrm{img}}(i;D)-\pi_{\mathrm{img}}(j;D).
\end{aligned}
\]
An intuitive illustration is shown in \cref{fig:method}.

\subsection{When \& Why Priors are Informative}

\paragraph{In a nutshell.}
(i) Fine-tuning mainly reshapes a low-dimensional “base” subspace spanned by base class prototypes; (ii) zero-shot geometry between classes is relatively stable across target domains; (iii) fine-tuning on base classes warps the pretrained inter-class directions, so our neutral priors become unreliable on base classes but remain informative for novel classes.

\paragraph{Preparation.}
We model fine-tuning as a low-rank deformation \cite{lora1, lora2, lora3} of the \emph{raw} pretrained geometry:
\begin{equation}
\label{eq:low-rank-raw}
\begin{aligned}
g_{\mathrm{img}}(x)\;&=\;g_{\mathrm{img}}^{0}(x)\;+\;U_{\mathrm{img}}\,a(x),\\
g_{\mathrm{txt}}(z)\;&=\;g_{\mathrm{txt}}^{0}(z)\;+\;U_{\mathrm{txt}}\,b(z),
\end{aligned}
\end{equation}
with $U_{\mathrm{img}},U_{\mathrm{txt}}\in\mathbb{R}^{d\times r}$, $r\!\ll\! d$, and coefficient maps $a(\cdot),b(\cdot)$ that are \emph{large} on $\mathcal{B}$ and \emph{small on average} on $\mathcal{N}$. Normalized features are obtained by $f=\mathrm{norm}(g)$ and $t(\cdot)=\mathrm{norm}(g_{\mathrm{txt}}(\cdot))$.
We define the pretrained inter-class direction
\(
\Delta^{0}_{ij}:=t^{0}(i)-t^{0}(j)
\)
and, for any feature $v\in\mathbb{R}^d$, the (signed) separation
\(
\delta^{0}_{ij}(v):=\langle v,\,\Delta^{0}_{ij}\rangle.
\)
We write \(\Pi_T\) for the Euclidean orthogonal projector onto a subspace \(T\).
\begin{assumption}[Base subspace]
\label{assump:base-subspace}
Recall $S=\mathrm{span}\{t^{0}(c):c\in\mathcal{B}\}$. For most $(i,j)\in\mathcal{B}\times\mathcal{B}$,
\(
\|\Pi_{S^\perp}(t(i)-t(j))\|\ll\|\Pi_{S}(t(i)-t(j))\|.
\)
\end{assumption}
In practice, efficient fine-tuning of VLMs—using only a small number of trainable parameters or low-rank adapters—often matches full fine-tuning on the base classes $\mathcal{B}$, indicating that discriminative updates concentrate along a few shared directions (i.e., within ($S$)).

\begin{assumption}[Zero-shot stability]

\label{assump:zs-stability}
For $x\!\sim\!D$ labeled in $\mathcal{N}$ and typical confusable pairs $(i,j)\in \binom{\mathcal{N}}{2}$,
\[
\mathrm{Var}_{x\sim D}\big[\delta^{0}_{ij}(f^{0}_{\mathrm{img}}(x))\big]\le \sigma^2,
\]
with a small, domain-insensitive $\sigma^2$.
\end{assumption}
The assumption follows from a one-line Rayleigh-quotient argument under mild conditions, for example:
let $\|\Delta^{0}_{ij}\|=1$, and define $\Omega_D:=\mathrm{Cov}_{x\sim D}[f^{0}_{\mathrm{img}}(x)]$, then
\[
\mathrm{Var}_{x\sim D}\big[\delta^{0}_{ij}(f^{0}_{\mathrm{img}}(x))\big]
=(\Delta^{0}_{ij})^\top \Omega_D \Delta^{0}_{ij}
\le \lambda_{\max}(\Omega_D),
\]
where \(\lambda_{\max}(\Omega_D)\) denotes the largest eigenvalue of \(\Omega_D\). Hence, when specific conditions are met, Assumption~\ref{assump:zs-stability} holds as a lemma. This assumption is also consistent with modeling L2-normalized embeddings on the unit sphere via a von Mises–Fisher family \cite{vmf1, vmf2}.

\begin{lemma}[Gaps as projections]
\label{lem:prior-proj}
For $(i,j)\in\mathcal{N}\times\mathcal{N}$,
\[
\begin{aligned}
\Delta\pi_{\mathrm{img}}(i,j)
&= \langle u_{\mathrm{img}}(D),\,\Delta^{0}_{ij}\rangle,\\
\Delta\pi_{\mathrm{txt}}(i,j)
&= \langle \Delta^{0}_{ij},\,u_{\mathrm{txt}}^{0}(D)\rangle\\
&\quad + \bigl\langle U_{\mathrm{txt}}\bigl(b(i)-b(j)\bigr),\,u_{\mathrm{txt}}^{0}(D)\bigr\rangle .
\end{aligned}
\]
In particular,
\[
\begin{aligned}
& \big|\Delta\pi_{\mathrm{txt}}(i,j)-\langle \Delta^{0}_{ij},u_{\mathrm{txt}}^{0}(D)\rangle\big|
\; \\ &\lesssim\; \|U_{\mathrm{txt}}\|\big(\|b(i)\|+\|b(j)\|\big),
\end{aligned}
\]
\(\lesssim\) denotes inequality up to an absolute constant factor. So both priors probe the same pretrained direction $\Delta^{0}_{ij}$ up to a small novel-class residual.
\end{lemma}

\noindent\textit{Takeaway.} The image prior uses a mean-image feature anchor; the text prior uses a neutral text anchor. Both are linear “rulers” aligned with pretrained inter-class directions on novel classes.

\begin{lemma}[Base suppression]
\label{lem:base-suppress}
Conditional on Assumption~\ref{assump:base-subspace}, if
\(
\|\Pi_S u_{\mathrm{img}}(D)\|\le \kappa_{\mathrm{img}}
\)
and
\(
\|\Pi_S u_{\mathrm{txt}}^{0}(D)\|\le \kappa_{\mathrm{txt}},
\)
then for any $(i,j)\in\mathcal{B}\times\mathcal{B}$,
\[
\begin{aligned}
|\Delta\pi_{\mathrm{img}}(i,j)|
&\le \kappa_{\mathrm{img}}\|\Pi_S \Delta^{0}_{ij}\|,\\
|\Delta\pi_{\mathrm{txt}}(i,j)|
&\le \kappa_{\mathrm{txt}}\|\Pi_S(t(i)-t(j))\|\\
&\quad + \|\Pi_{S^\perp}u_{\mathrm{txt}}^{0}(D)\|\,
         \|\Pi_{S^\perp}(t(i)-t(j))\|.
\end{aligned}
\]
\end{lemma}

\noindent\textit{Takeaway.} Because base directions live in $S$ and the neutral anchors have little energy in $S$, neutral-prior gaps are intrinsically small on $\mathcal{B}$—yet remain informative off $S$ for novel classes.

\begin{figure}[t]
  \centering
   \includegraphics[width=0.95\linewidth]{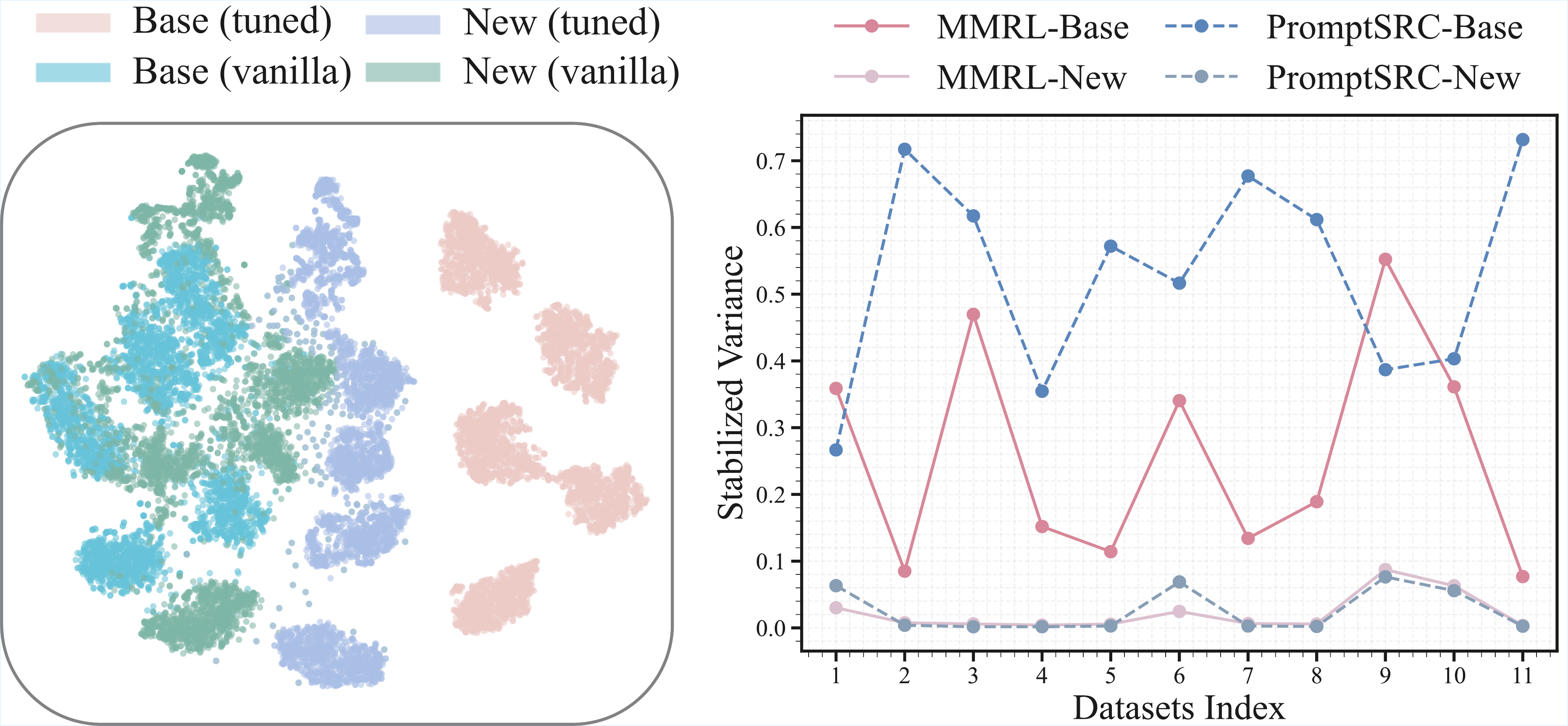}
   \caption{The t-SNE visualization of embeddings, and the comparison of between-class variance of per-class mean logits.}
   \label{fig:tsne}
   \vspace{-0.4cm}
\end{figure}

We have also visualized the spatial distribution of embeddings on base and new classes using t-SNE, as well as the stabilized variance of the output logits with neutral-reference images as input on different datasets as \cref{fig:tsne} shows.

For a pair $(i,j)$, define the logits margin:
\[
m_{ij}(x)\;:=\;\ell_i(x)-\ell_j(x)\;=\;\big\langle f_{\mathrm{img}}(x),\,t(i)-t(j)\big\rangle,
\]
and its expectation $\mu_{ij}(D):=\mathbb{E}_{x\sim D}[m_{ij}(x)]$. 
By combining Assumption~\ref{assump:base-subspace}, Assumption~\ref{assump:zs-stability}, Lemma~\ref{lem:prior-proj} and Lemma~\ref{lem:base-suppress}, we can explain why priors are informative.

\begin{proposition}[Sign consistency]
\label{prop:prior-alignment}
Under the low-rank deformation model in~\eqref{eq:low-rank-raw} and Assumption~\ref{assump:zs-stability}, there exist domain- and model-dependent constants
\[
\varepsilon_{\mathrm{img}}^{\mathcal{N}},\ \varepsilon_{\mathrm{txt}}^{\mathcal{N}}\;\;\text{(small on $\mathcal{N}$)}
\]
such that for any $(i,j)\in\mathcal{N}\times\mathcal{N}$,
\[
\begin{aligned}
\bigl|\mu_{ij}(D)-\Delta\pi_{\mathrm{img}}(i,j)\bigr|
&\le \varepsilon_{\mathrm{img}}^{\mathcal{N}},\\
\bigl|\mu_{ij}(D)-\Delta\pi_{\mathrm{txt}}(i,j)\bigr|
&\le \varepsilon_{\mathrm{txt}}^{\mathcal{N}}.
\end{aligned}
\]
Consequently, $\operatorname{sign}\,\mu_{ij}(D)=\operatorname{sign}\,\Delta\pi_{\mathrm{img}}(i,j)$ and
$\operatorname{sign}\,\mu_{ij}(D)=\operatorname{sign}\,\Delta\pi_{\mathrm{txt}}(i,j)$ whenever the gaps exceed the respective $\varepsilon$.
\end{proposition}

On novel classes, because the inter-class direction satisfies $t(i)-t(j)\approx \Delta^{0}_{ij}$ and the low-rank deformation implies $f_{\mathrm{img}}(x)-f_{\mathrm{img}}^{0}(x)$ is small on $\mathcal{N}$, thus we have $\mathrm{Var}[m_{ij}(x)]\le \sigma^2+o(1)$, i.e., $\sigma_m^2\lesssim \sigma^2$.

\begin{corollary}[High-probability sign consistency]
\label{cor:sign-consistency}
Let $(i,j)\in\mathcal{N}\times\mathcal{N}$ and write $\mu:=\mu_{ij}(D)$ and
$\sigma_m^2:=\mathrm{Var}_{x\sim D}\!\big[m_{ij}(x)\big]$. Moreover, on $\mathcal{N}$ we have $\sigma_m^2 \lesssim \sigma^2$.
Then for any $\gamma>0$,
\[
\mathbb{P}_{x\sim D}\Big[\operatorname{sign} m_{ij}(x)\neq \operatorname{sign}\mu\Big]
\le
\min\left\{\frac{\sigma_m^{2}}{(\lvert\mu\rvert-\gamma)^{2}},1\right\},
\]
whenever $\lvert\mu\rvert>\gamma$.
In particular, if $|\Delta\pi_{\mathrm{img}}(i,j)|>\varepsilon_{\mathrm{img}}^{\mathcal{N}}+\gamma$
(or $|\Delta\pi_{\mathrm{txt}}(i,j)|>\varepsilon_{\mathrm{txt}}^{\mathcal{N}}+\gamma$),
then with probability at least
$1-\sigma_m^2/(\,|\Delta\pi|-\varepsilon-\gamma\,)^2$ the sample-level sign agrees with the prior.
\end{corollary}

\noindent\textit{Interpretation.} For novel classes, both priors measure the zero-shot inter-class bias direction using neutral anchors. Low-rank fine-tuning perturbs this geometry weakly on $\mathcal{N}$, so the \emph{expected} logits gap aligns with the prior gap up to a small error; the zero-shot stability assumption then upgrades this to a \emph{high-probability} statement at the sample level.

Formal proof and more details are in the Appendix.

\subsection{Bayes-Style Surrogate}

\paragraph{Neutral-prior composition.}
For a domain $D$ and class $c\in\mathcal{C}$, recall
\[
\label{eq:prior-txt-impl-fixed}
\pi_{\mathrm{txt}}(c;D)
= \big\langle t(c),\,u_{\mathrm{txt}}^{0}(D)\big\rangle.
\]
On the image side, recall $u_{\mathrm{img}}(D)\;:=\;f_{\mathrm{img}}(\bar x^{D})$ and
\[
\label{eq:prior-img-impl-fixed}
\pi_{\mathrm{img}}(c;D)
= \big\langle u_{\mathrm{img}}(D),\,t^{0}(c)\big\rangle.
\]


Given a pair $(i,j)$, we form a composite neutral prior gap
\begin{equation}
\label{eq:Sigma-impl}
\begin{aligned}
\Sigma_{i,j}(D)
:= &\big(\pi_{\mathrm{txt}}(i;D)-\pi_{\mathrm{txt}}(j;D)\big) \\
 \;&+\; \big(\pi_{\mathrm{img}}(i;D)-\pi_{\mathrm{img}}(j;D)\big).
\end{aligned}
\end{equation}
Note $\Sigma_{i,j}(D)=-\Sigma_{j,i}(D)$. We restrict comparisons to a symmetric neighborhood graph $G\subset\mathcal{C}\times\mathcal{C}$ (i.e., $(i,j)\in G \Leftrightarrow (j,i)\in G$) and write $\mathcal{A}(i)=\{j\neq i:(i,j)\in G\}$. To absorb domain-wide constant bias caused by fine-tuning in the composite neutral prior, we use a global intercept on the base pairs:
\begin{equation}
\label{eq:global-intercept-def}
\hat \beta(D):=\operatorname*{argmin}_{\beta}
\sum_{(i,j)\in\mathcal{B}\times\mathcal{B}}
\big(\hat\mu_{ij}(D)-\Sigma_{i,j}(D)-\beta\big)^2,
\end{equation}
whose closed-form solution $\beta^\star(D)$ and $\hat\mu_{ij}(D)$ is the empirical estimated domain mean of $m_{ij}(x)$, details of both are given in the Appendix. To reduce complexity, we combine it with $\tau$ in \cref{eq:prior-dominated-region-fixed} in practice.

\paragraph{Residual form for priors.}

Considering the diverse styles of downstream data, we provide a residual form to better adapt to semantically diverse datasets (e.g., ImageNet). On such datasets, inter-class semantics and text prototypes are more widely separated; combining Assumption~\ref{assump:zs-stability}, the across-pair variance of zero-shot projections (and thus logits gaps) becomes larger. To stabilize the neutral priors and cancel common, domain/prompt–induced bias, we \emph{residualize} the image- or text-side prior by subtracting the correlation with the \emph{current} anchor from that with the \emph{zero-shot} anchor—thereby measuring correlation with the anchor displacement and retaining only domain-induced changes.

Let the current-model text anchor under the neutral prompt $\tau(D)$ be
$
u_{\mathrm{txt}}(D):=\operatorname{norm}\!\big(g_{\mathrm{txt}}(\tau(D))\big).
$
We define the \emph{text residual prior}
\begin{equation}
\label{eq:residual-text-prior}
\tilde{\pi}_{\mathrm{txt}}(c;D)
:= {\langle t(c),\,u_{\mathrm{txt}}^{0}(D)\rangle}
- {\langle t(c),\,u_{\mathrm{txt}}(D)\rangle} .
\end{equation}
Similarly, using $u_{\mathrm{img}}(D)$ and class prototypes, we define the \emph{image residual prior}
\begin{equation}
\label{eq:residual-img-prior}
\tilde{\pi}_{\mathrm{img}}(c;D)
:= {\langle u_{\mathrm{img}}(D),\,t^{0}(c)\rangle}
- {\langle u_{\mathrm{img}}(D),\,t(c)\rangle} .
\end{equation}
The pairwise gaps become
\[
\begin{aligned}
\Delta\tilde{\pi}_{\mathrm{txt}}(i,j)&=\Delta\pi_{\mathrm{txt}}(i,j)-\langle t(i)-t(j),u_{\mathrm{txt}}(D)\rangle,\\
\Delta\tilde{\pi}_{\mathrm{img}}(i,j)&=\Delta\pi_{\mathrm{img}}(i,j)-\langle u_{\mathrm{img}}(D),t(i)-t(j)\rangle,
\end{aligned}
\]
Intuitively, $\Delta\tilde{\pi}_{\mathrm{txt}}(i,j)\approx\langle \Delta^{0}_{ij},\,u_{\mathrm{txt}}^{0}(D)-u_{\mathrm{txt}}(D)\rangle$ projects the pretrained inter-class direction onto the anchor displacement, filtering out class-independent bias and reducing across-pair variance on semantically diverse datasets; $\Delta\tilde{\pi}_{\mathrm{img}}(i,j)$ analogously measures deformation of inter-class directions and is typically small on novel classes.

We then use
\begin{equation}
\label{eq:residual-sigma}
\begin{aligned}
\tilde{\Sigma}_{i,j}(D)
:= &\big(\tilde{\pi}_{\mathrm{txt}}(i;D)-\tilde{\pi}_{\mathrm{txt}}(j;D)\big)\\
&+\big(\tilde{\pi}_{\mathrm{img}}(i;D)-\tilde{\pi}_{\mathrm{img}}(j;D)\big)
\end{aligned}
\end{equation}
as a drop-in replacement for $\Sigma_{i,j}(D)$ in \eqref{eq:Sigma-impl}. This replacement preserves the sign-consistency guarantee (Proposition~\ref{prop:prior-alignment}) and the high-probability of this consistency (Corollary~\ref{cor:sign-consistency}), often with tighter constants due to variance reduction; full details are deferred to the Appendix.

\begin{figure}[t]
  \centering
   \includegraphics[width=0.95\linewidth]{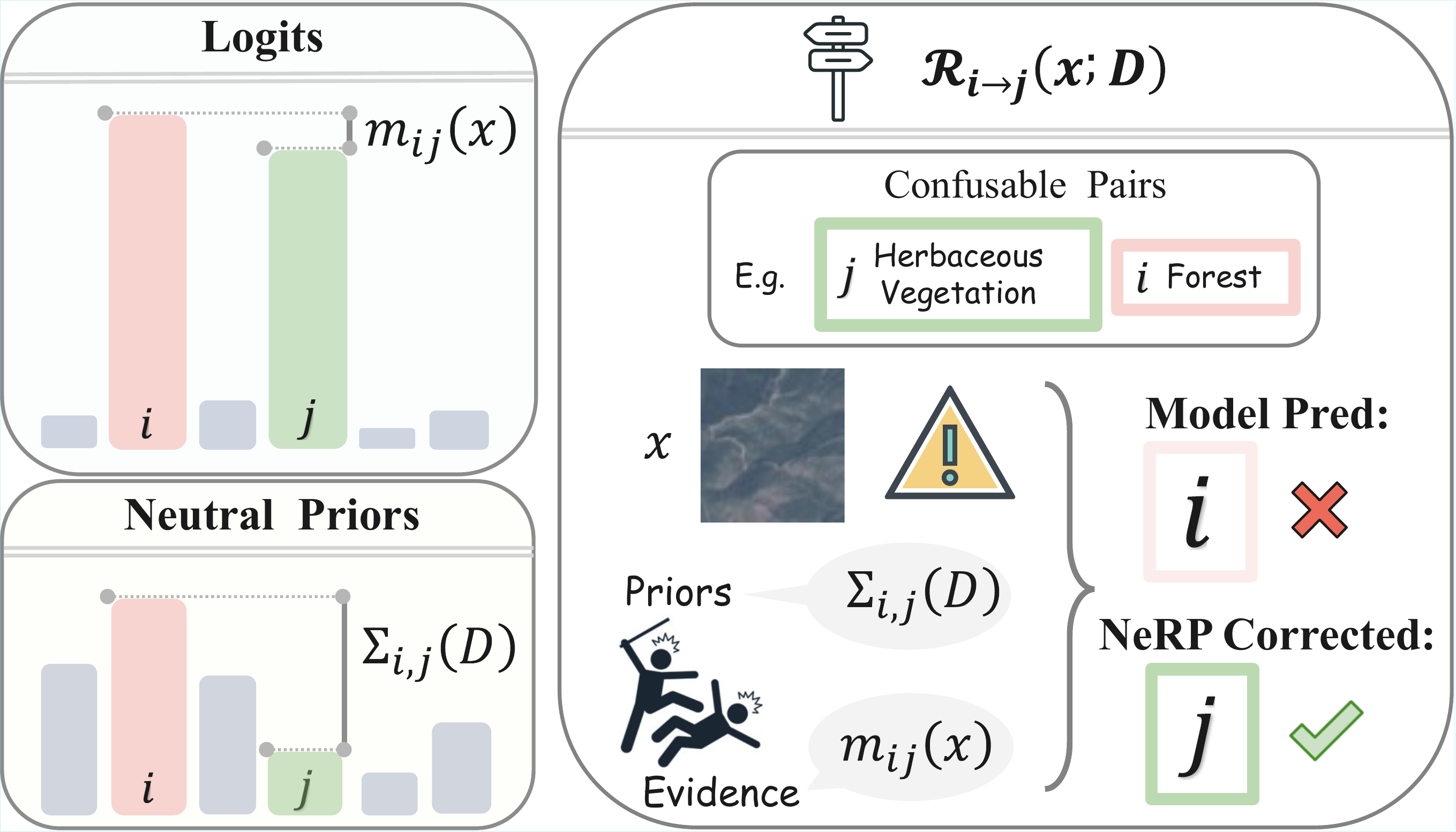}
   \caption{An explanation of how NeRP corrects mispredictions.}
   \label{fig:flip}
   \vspace{-0.4cm}
\end{figure}

\vspace{-0.3cm}
\paragraph{Prior-dominated inconsistency.}
For a test image $x$ with top-1 prediction $\hat y(x)=i$, and any $j\in\mathcal{A}(i)$, recall the sample-level margin $m_{ij}(x)=\ell_i(x)-\ell_j(x)$. We interpret $\Sigma_{i,j}(D)$ as a prior logit-style offset for $i$ against $j$, and $m_{ij}(x)$ as the data-dependent logit margin. \(m_{ij}(x)\) comes from L2-normalized cosine similarities between embeddings, and under a von Mises–Fisher model, this margin can be approximately interpreted as a log-likelihood ratio. A Bayes-inspired surrogate score is then

\begin{equation}
\label{eq:posterior-surrogate-fixed}
s_{ij}(x)\;\approx\; m_{ij}(x)\;+\;\Sigma_{i,j}(D)\;+\;\hat \beta(D).
\end{equation}

By Proposition~\ref{prop:prior-alignment} and additivity of errors, on novel classes there exists $\varepsilon^{\mathcal{N}}=\varepsilon_{\mathrm{img}}^{\mathcal{N}}+\varepsilon_{\mathrm{txt}}^{\mathcal{N}}$ such that
\begin{equation}
\label{eq:composite-alignment-fixed}
\bigl|\mu_{ij}(D)-\Sigma_{i,j}(D)\bigr|\ \le\ \varepsilon^{\mathcal{N}}.
\end{equation}
Thus, if the prior strongly favors $i$ against $j$ (i.e., $\Sigma_{i,j}(D)$ is large), we want the \emph{expected} margin $\mu_{ij}(D)$ to be comparably large as well. When, however, the \emph{observed} margin $m_{ij}(x)$ is small, the posterior surrogate in~\eqref{eq:posterior-surrogate-fixed} is driven mainly by the prior term rather than by sample evidence.
We then declare a prior-dominated region by prior gate $\tau$ and evidence gate $\delta$:
\begin{equation}
\label{eq:prior-dominated-region-fixed}
\mathcal{R}_{i\to j}(x;D) :=
\Big\{\Sigma_{i,j}(D) \ge {\tau-\hat \beta(D)} \wedge m_{ij}(x)\le\delta \Big\}.
\end{equation}
Inside $\mathcal{R}_{i\to j}$, the gap $\Sigma_{i,j}(D)-m_{ij}(x)\ge \tau-\delta-\hat \beta(D)$ indicates the surrogate posterior is predominantly prior-driven.

\begin{table*}[!t]
\centering
\setlength{\abovecaptionskip}{0.15cm}   
\caption{Base-to-novel generalization experiments of five baselines on 11 datasets. “\bigdot” denotes keeping the original output. }
\label{tab:base_to_novel}
\renewcommand\arraystretch{0.95}
\setlength{\tabcolsep}{2.9mm}{
\resizebox{0.9\textwidth}{!}{
    \begin{tabular}{@{}>{\centering\arraybackslash}p{3.2cm}ccc|ccc|ccc|ccc@{}}
    \toprule[1.5pt]
    \multirow{2}{*}{Method} &
      \multicolumn{3}{c}{Average} &
      \multicolumn{3}{c}{ImageNet} &
      \multicolumn{3}{c}{Caltech101} &
      \multicolumn{3}{c}{OxfordPets} \\
     &
      Base & Novel & HM & Base & Novel & HM &
      Base & Novel & HM & Base & Novel & HM \\ \midrule
    CoOp~\scriptsize{\textcolor{gray}{(IJCV 22)}} &
      82.69 & 63.22 & 71.66 & 76.47 & 67.88 & 71.92 &
      98.00 & 89.81 & 93.73 & 93.67 & 95.29 & 94.47 \\
    MaPLe~\scriptsize{\textcolor{gray}{(CVPR 23)}} &
      82.28 & 75.14 & 78.55 & 76.66 & 70.54 & 73.47 &
      97.74 & 94.36 & 96.02 & 95.43 & 97.76 & 96.58 \\
    CoPrompt~\scriptsize{\textcolor{gray}{(ICLR 24)}} &
      84.00 & 77.23 & 80.48 & 77.67 & 71.27 & 74.33 &
      98.27 & 94.90 & 96.55 & 95.67 & 98.10 & 96.87 \\
    SkipT~\scriptsize{\textcolor{gray}{(CVPR 25)}} &
      85.04 & 77.53 & 81.11 & 77.73 & 70.40 & 73.89 &
      98.50 & 95.33 & 96.89 & 95.70 & 97.87 & 96.77
      \\ \midrule
    CoCoOp~\scriptsize{\textcolor{gray}{(CVPR 22)}} &
      80.47 & 71.69 & 75.83 & 75.98 & 70.43 & 73.10 &
      97.96 & 93.81 & 95.84 & 95.20 & 97.69 & 96.43 \\
    +NeRP~\scriptsize{\textcolor{cvprblue}{(Ours)}} &
      \bigdot & \textbf{72.99} & \textbf{76.67} &
      \bigdot & 71.37 & 73.60 &
      \bigdot & 94.07 & 95.98 &
      \bigdot & 97.90 & 96.53 \\
    PromptSRC~\scriptsize{\textcolor{gray}{(ICCV 23)}} &
      84.26 & 76.10 & 79.97 & 77.60 & 70.73 & 74.01 &
      98.10 & 94.03 & 96.02 & 95.33 & 97.30 & 96.30 \\
    +NeRP~\scriptsize{\textcolor{cvprblue}{(Ours)}} &
      \bigdot & \textbf{76.87} & \textbf{80.40} &
      \bigdot & 71.33 & 74.33 &
      \bigdot & 94.47 & 96.25 &
      \bigdot & 97.40 & 96.35 \\
    MMA~\scriptsize{\textcolor{gray}{(CVPR 24)}} &
      83.20 & 76.80 & 79.87 & 77.31 & 71.00 & 74.02 &
      98.40 & 94.00 & 96.15 & 95.40 & 98.07 & 96.72 \\
    +NeRP~\scriptsize{\textcolor{cvprblue}{(Ours)}} &
      \bigdot & \textbf{77.59} & \textbf{80.30} &
      \bigdot & 71.73 & 74.42 &
      \bigdot & 94.27 & 96.29 &
      \bigdot & 98.13 & 96.75 \\
    MMRL~\scriptsize{\textcolor{gray}{(CVPR 25)}} &
      85.68 & 77.16 & 81.20 & 77.90 & 71.30 & 74.45 &
      98.97 & 94.50 & 96.68 & 95.90 & 97.60 & 96.74 \\ 
    +NeRP~\scriptsize{\textcolor{cvprblue}{(Ours)}} &
      \bigdot & \textbf{78.04} & \textbf{81.68} &
      \bigdot & 71.53 & 74.58 &
      \bigdot & 95.10 & 97.00 &
      \bigdot &  97.70 & 96.79 \\\midrule \midrule
    \multirow{2}{*}{Method} &
      \multicolumn{3}{c}{StanfordCars} &
      \multicolumn{3}{c}{Flowers102} &
      \multicolumn{3}{c}{Food101} &
      \multicolumn{3}{c}{FGVCAircraft} \\
     &
      Base & Novel & HM & Base & Novel & HM &
      Base & Novel & HM & Base & Novel & HM \\ \midrule
    CoOp~\scriptsize{\textcolor{gray}{(IJCV 22)}} &
      78.12 & 60.40 & 68.13 & 97.60 & 59.67 & 74.06 &
      88.33 & 82.26 & 85.19 &  40.44 & 22.30 & 28.75 \\
    MaPLe~\scriptsize{\textcolor{gray}{(CVPR 23)}} &
      72.94 & 74.00 & 73.47 & 95.92 & 72.46 & 82.56 &
      90.71 & 92.05 & 91.38 & 37.44 & 35.61 & 36.50 \\
    CoPrompt~\scriptsize{\textcolor{gray}{(ICLR 24)}} &
      76.97 & 74.40 & 75.66 & 97.27 & 76.60 & 85.71 &
      90.73 & 92.07 & 91.40 & 40.20 & 39.33 & 39.76 \\

    SkipT~\scriptsize{\textcolor{gray}{(CVPR 25)}} &
      82.93 &  72.50 & 77.37 & 98.57 & 75.80 &  85.70 &
      90.67 & 92.03 &  91.34 & 45.37 & 37.13 & 40.84
      \\ \midrule
    CoCoOp~\scriptsize{\textcolor{gray}{(CVPR 22)}} &
      70.49 & 73.59 & 72.01 & 94.87 & 71.75 & 81.71 &
      90.70 & 91.29 & 90.99 & 33.41 & 23.71 & 27.74 \\
    +NeRP~\scriptsize{\textcolor{cvprblue}{(Ours)}} &
      \bigdot & 74.13 & 72.26 &
      \bigdot & 72.03 & 81.89 &
      \bigdot & 91.33 &  91.01 &
      \bigdot & 26.83 & 29.76 \\
    PromptSRC~\scriptsize{\textcolor{gray}{(ICCV 23)}} &
      78.27 & 74.97 & 76.58 & 98.07 & 76.50 & 85.95 &
      90.67 & 91.53 & 91.10 & 42.73 & 37.87 &  40.15 \\
    +NeRP~\scriptsize{\textcolor{cvprblue}{(Ours)}} &
      \bigdot & 75.70 &  76.96 &
      \bigdot & 77.37 & 86.50 &
      \bigdot &  91.53 & 91.10 &
      \bigdot & 38.07 & 40.27 \\
    MMA~\scriptsize{\textcolor{gray}{(CVPR 24)}} &
      78.50 & 73.10 &  75.70 & 97.77 & 75.93 & 85.48 &
      90.13 & 91.30 & 90.71 & 40.57 & 36.33 & 38.33 \\
    +NeRP~\scriptsize{\textcolor{cvprblue}{(Ours)}} &
      \bigdot & 73.97 & 76.17 &
      \bigdot & 77.03 &  96.17 &
      \bigdot &  91.30 &  90.71 &
      \bigdot &  36.63 & 38.50 \\
    MMRL~\scriptsize{\textcolor{gray}{(CVPR 25)}} &
      81.30 &  75.07 & 78.06 & 98.97 &  77.27 &  86.78 &
      90.57 & 91.50 & 91.03 & 46.30 & 37.03 & 41.15 \\ 
    +NeRP~\scriptsize{\textcolor{cvprblue}{(Ours)}} &
      \bigdot & 75.30 & 78.19 &
      \bigdot & 78.43 & 87.51 &
      \bigdot & 91.55 & 91.06 &
      \bigdot & 37.67 & 41.54 \\ \midrule \midrule
    \multirow{2}{*}{Method} &
      \multicolumn{3}{c}{SUN397} &
      \multicolumn{3}{c}{DTD} &
      \multicolumn{3}{c}{EuroSAT} &
      \multicolumn{3}{c}{UCF101} \\
     &
      Base & Novel & HM & Base & Novel & HM &
      Base & Novel & HM & Base & Novel & HM \\ \midrule
    CoOp~\scriptsize{\textcolor{gray}{(IJCV 22)}} &
      80.60 & 65.89 & 72.51 & 79.44 & 41.18 & 54.24 &
      92.19 & 54.74 & 68.69 & 84.69 & 56.05 & 67.46 \\
    MaPLe~\scriptsize{\textcolor{gray}{(CVPR 23)}} &
      80.82 & 78.70 & 79.75 &  80.36 & 59.18 & 68.16 &
      94.07 & 73.23 & 82.35 & 83.00 &  78.66 & 80.77 \\
    CoPrompt~\scriptsize{\textcolor{gray}{(ICLR 24)}} &
      82.63 & 80.03 & 81.30 & 83.13 & 64.73 & 72.79 &
      94.60 & 78.57 & 85.84 & 86.90 & 79.57 & 83.07 \\
    SkipT~\scriptsize{\textcolor{gray}{(CVPR 25)}} &
      82.40 &  79.03 & 80.68 & 83.77 & 67.23 & 74.59 &
      92.47 & 83.00 & 87.48 & 87.30 & 82.47 & 84.81
      \\ \midrule
    CoCoOp~\scriptsize{\textcolor{gray}{(CVPR 22)}} &
      79.74 & 76.86 & 78.27 & 77.01 & 56.00 & 64.85 &
      87.49 & 60.04 & 71.21 & 82.33 & 73.45 & 77.64 \\
    +NeRP~\scriptsize{\textcolor{cvprblue}{(Ours)}} &
      \bigdot & 77.46 & 78.58 &
      \bigdot & 56.67 & 65.29 &
      \bigdot & 67.30 & 76.08 &
      \bigdot & 73.77 & 77.82 \\
    PromptSRC~\scriptsize{\textcolor{gray}{(ICCV 23)}} &
      82.67 & 78.47 & 80.52 & 83.37 & 62.97 & 71.75 &
      92.90 & 73.90 & 82.32 & 87.10 & 78.80 & 82.74 \\
    +NeRP~\scriptsize{\textcolor{cvprblue}{(Ours)}} &
      \bigdot & 78.90 & 80.74 &
      \bigdot & 63.37 & 72.01 &
      \bigdot & 78.53 & 85.11 &
      \bigdot & 78.93 & 82.81 \\
    MMA~\scriptsize{\textcolor{gray}{(CVPR 24)}} &
      82.27 & 78.57 & 80.38 & 83.20 & 65.63 & 73.38 &
      85.46 & 82.34 &  83.87 & 86.23 & 80.03 & 82.20 \\
    +NeRP~\scriptsize{\textcolor{cvprblue}{(Ours)}} &
      \bigdot &  78.77 &  80.48 &
      \bigdot &  66.20 & 73.73 &
      \bigdot & 85.27 &  85.36 &
      \bigdot &  80.17 & 83.09 \\
    MMRL~\scriptsize{\textcolor{gray}{(CVPR 25)}} &
      83.20 & 79.30 & 81.20 & 85.67 & 65.00 &  73.82 &
      95.60 &  80.17 &  87.21 & 88.10 & 80.07 &  83.89 \\ 
    +NeRP~\scriptsize{\textcolor{cvprblue}{(Ours)}} &
      \bigdot &  79.60 & 81.36 &
      \bigdot & 65.70 & 74.37 &
      \bigdot & 85.17 & 90.08 &
      \bigdot & 80.67 & 84.22 \\ \bottomrule[1.5pt]
    \end{tabular}
            }
}
\vspace{-0.4cm}
\end{table*}

\begin{table*}[ht]
    \centering
    \setlength{\abovecaptionskip}{0.15cm}   
    \caption{Cross-dataset generalization experiments on 10 datasets.}
    \label{tab:cross_dataset}
    \renewcommand\arraystretch{0.9}
    \setlength{\tabcolsep}{2.0mm}{
    \resizebox{0.88\linewidth}{!}
    {
        \begin{tabular}{ccccccccccccccc}
        \toprule[1.5pt]
        \multirow{3}[3]{*}{Method} & Source & \multicolumn{10}{c}{Target Dataset} & \multirow{3}[3]{*}{Average} \\
        \cmidrule(lr){2-2}\cmidrule(lr){3-12}
         ~ & Image & Caltech & Oxford & Stanford & Flowers & \multirow{2}*{Food101} & FGVC & \multirow{2}*{SUN397} & \multirow{2}*{DTD} & Euro & \multirow{2}*{UCF101} &  \\
        ~ & Net    & 101     & Pets   & Cars      & 102     & ~ & Aircraft & ~ & ~ & SAT  & ~ & ~ \\
        \midrule
        CoCoOp & 71.02 & 94.43 & 90.14 & 65.32 & 71.88 & 86.06 & 22.94 & 67.36 & 45.73 & 45.37 & 68.21 & 65.74 \\
        +NeRP~\scriptsize{\textcolor{cvprblue}{(Ours)}} & \bigdot & 94.87 & 90.73 & 65.35 & 72.53 & 86.30 & 23.70 & 67.43 & 46.70 & 51.20 & 68.97 & \textbf{66.78} \\
        \midrule
        PromptSRC & 71.27 & 93.60 & 90.25 & 65.70 & 70.25 & 86.15 & 23.90 & 67.10 & 46.87 & 45.50 & 68.75 & 65.81\\
        +NeRP~\scriptsize{\textcolor{cvprblue}{(Ours)}} & \bigdot & 94.10 & 90.30 & 66.20 & 71.37 & 86.40 & 24.23 & 67.30 & 47.83 & 51.70 & 69.70 & \textbf{66.91}  \\
        \midrule
        MMA & 71.00 & 93.80 & 90.30 & 66.13 & 72.07 & 86.12 & 25.33 & 68.17 & 46.57 & 49.24 & 68.32 & 66.61 \\
        +NeRP~\scriptsize{\textcolor{cvprblue}{(Ours)}} & \bigdot & 94.27 & 90.47 & 66.53 & 73.10 & 86.12 & 25.87 & 68.25 & 47.03 & 53.50 & 69.80 & \textbf{67.49} \\
        \midrule
        MMRL & 72.03 & 94.67 & 91.43 & 66.10 & 72.77 & 86.40 & 26.30 & 67.57 & 45.90 & 53.10 & 68.27 & 67.25\\
        +NeRP~\scriptsize{\textcolor{cvprblue}{(Ours)}} & \bigdot & 94.77 & 91.43 & 66.60 & 73.97 & 86.50 & 26.50 & 67.73 & 46.17 & 55.87 & 70.43 & \textbf{68.00}  \\
        \bottomrule[1.5pt]
        \end{tabular}
    }}
    \vspace{-0.2cm}
\end{table*}

\begin{table}[ht]
    \centering
    \setlength{\abovecaptionskip}{0.15cm}   
    \caption{Domain generalization experiments on 4 datasets.
    }
    \label{tab:domain_dataset}
    \resizebox{0.9\linewidth}{!}
    {
        \begin{tabular}{ccccccc}
        \toprule
        \multirow{2}[2]{*}{Method} & Source & \multicolumn{4}{c}{Target Dataset} & \multirow{2}[2]{*}{Average} \\
        \cmidrule(lr){2-2}\cmidrule(lr){3-6}
         ~  & ImageNet & -V2 & -S & -A & -R & ~ \\
         \midrule
        CoCoOp & 71.02 & 64.07 & 48.75 & 50.63 & 76.18 & 59.91 \\
        +NeRP~\scriptsize{\textcolor{cvprblue}{(Ours)}} & \bigdot & 66.73 & 50.20 & 52.47 & 77.00 & \textbf{61.60} \\
        \midrule
        PromptSRC & 71.27 & 64.35 & 49.55 & 50.90 & 77.80 & 60.65 \\
        +NeRP~\scriptsize{\textcolor{cvprblue}{(Ours)}} & \bigdot & 66.87 & 49.77 & 52.77 & 78.13 & \textbf{61.89} \\
        \midrule
        MMA & 71.00 & 64.33 & 49.13 & 51.12 & 77.32 & 60.48 \\
        +NeRP~\scriptsize{\textcolor{cvprblue}{(Ours)}} & \bigdot & 67.77 & 50.07 & 51.77 & 78.20 & \textbf{61.95} \\
        \midrule
        MMRL  & 72.03 & 64.47 & 49.17 & 51.12 & 77.32 & 60.52 \\
        +NeRP~\scriptsize{\textcolor{cvprblue}{(Ours)}} & \bigdot & 66.63 & 49.27 & 53.33 & 77.93 & \textbf{61.79} \\
        \bottomrule
        \end{tabular}
    }
    \vspace{-0.4cm}
\end{table}

\vspace{-0.3cm}
\paragraph{Decision rule.}
As \cref{fig:flip} shows, for each $j\in\mathcal{A}(i)$, keep $j$ iff $(x,D)\in\mathcal{R}_{i\to j}$. Among retained neighbors, select
\[
j^\star \in \arg\max_{j:\,(x,D)\in\mathcal{R}_{i\to j}}\ \ell_j(x),
\]
and flip by enforcing a minimal tie-break on logits, e.g.,
\[
\ell_{j^\star}(x)\leftarrow \max\{\ell_{j^\star}(x),\,\ell_i(x)+\varepsilon_0\}\quad(\varepsilon_0>0).
\]
If no neighbor passes, leave logits unchanged.


\vspace{-0.3cm}
\paragraph{Risk--Benefit Analysis.}
Recall $\beta^\star(D)$ from \cref{eq:global-intercept-def}, and denote $\Delta_\beta:=\hat\beta(D)-\beta^\star(D)$. By Prop.~\ref{prop:prior-alignment} and error additivity, there exists 
\begin{equation}\label{eq:int-align}
\begin{gathered}
\varepsilon_{\mathrm{int}}^{\mathcal N} := \varepsilon^{\mathcal N} + \lvert \Delta_\beta \rvert,\\
\text{s.t.}\quad \bigl|\,\mu_{ij}(D) - \bigl(\Sigma_{i,j}(D) + \hat{\beta}(D)\bigr)\,\bigr|
\le \varepsilon_{\mathrm{int}}^{\mathcal N}.
\end{gathered}
\end{equation}

Define the \emph{effective gate} $\tilde\tau:=\tau-\hat\beta(D)$, and let $\sigma_m^2:=\mathrm{Var}_{x\sim D}[m_{ij}(x)]\lesssim\sigma^2$ under Assumption~\ref{assump:zs-stability}. On the event $\{\Sigma_{i,j} \ge \tilde\tau\}$, \cref{eq:int-align} implies $\mu_{ij}(D)\ge \tau-\varepsilon_{\mathrm{int}}^{\mathcal N}$. For any $\delta$ with
\[
\gamma\ :=\ \tau-\varepsilon_{\mathrm{int}}^{\mathcal N}-\delta\ >\ 0,
\]
Chebyshev yields
\[
\mathbb{P}_{x\sim D}\!\big[m_{ij}(x)\le\delta\big]\ \le\ \frac{\sigma_m^2}{\gamma^2}
\ =\ \frac{\sigma_m^2}{\big(\tau-\varepsilon_{\mathrm{int}}^{\mathcal N}-\delta\big)^2}.
\]
Because a flip can only occur when $m_{ij}(x)\le\delta$, the above equation upper-bounds the \emph{false-flip} probability. Importantly, this requires only
\(
\tau\ >\ \varepsilon_{\mathrm{int}}^{\mathcal N}+\delta.
\label{eq:tau-cond}
\)
In practice, we do not explicitly look for $\tau$ but directly seek a suitable $\tilde\tau$.

\section{Experiments}

\subsection{Settings}

\paragraph{Datasets \& Baselines}
Following previous works \cite{coop}, we evaluate base-to-new generalization and cross-dataset transfer on eleven recognition benchmarks spanning diverse data distributions: ImageNet \cite{imagenet}, Caltech101 \cite{caltech101}, OxfordPets \cite{oxford_pets}, StanfordCars \cite{stanford_cars}, Flowers102 \cite{flowers102}, Food101 \cite{food101}, FGVCAircraft \cite{fgvc_aircraft}, SUN397 \cite{sun397}, DTD \cite{dtd}, EuroSAT \cite{eurosat}, and UCF101 \cite{ucf101}. For cross-domain evaluation, we consider ImageNet-V2 \cite{imagenetv2}, ImageNet-Sketch \cite{imagenet_sketch}, ImageNet-A \cite{imagenet_a}, and ImageNet-R \cite{imagenet_r}. All experiments utilize a 16-shot setting (16 training samples per class), and accuracy is the average of 3 random seeds. 
We select 4 influential prompt learners (CoCoOp, PromptSRC, MMA and MMRL) with different styles as baselines for our plug-and-play strategy. We also include several state-of-the-art (SOTA) methods as reference baselines.


\vspace{-0.3cm}
\paragraph{Confusable Pairs Generation}
Let $\mathcal{C}$ denote the label set. For each class $i \in \mathcal{C}$, we query a locally deployed Qwen2.5-72B-Instruct \cite{qwen2} with the class name and description to produce at most $K$ candidate classes that are likely to be confused with $i$ (where $K$ scales with $|\mathcal{C}|$). Aggregating the resulting class pairs over all $i$, we construct an undirected adjacency graph $G$ on $\mathcal{C}$; an edge $(i,j)$ indicates that $j$ is a confusable neighbor of $i$. The prompt template is provided in the appendix.

\vspace{-0.3cm}
\paragraph{Target-Emulated Calibration}
NeRP’s decisions rely on two gates: the Prior Effective Gate \(\tilde{\tau}\) and the Evidence Strength Gate \(\delta\). Because the test data are unseen, we construct simulated subsets from the training data to select appropriate \(\tilde{\tau}\) and \(\delta\) for each task.

For the \textbf{base-to-novel} setting, we partition the base classes into \(e\) disjoint folds $\{F^{b2n}_e\}_{e=1}^E$. In fold \(e\), the pseudo-base classes are $\mathcal{B}_e=\mathcal{B}\setminus F_e$ and the pseudo-novel classes are \(\mathcal{N}_e=F_e\).
For the \textbf{cross-dataset setting}, following prior work, we group ImageNet-1K into superclasses, split them into \(e\) disjoint subsets $\{F^{cd}_e\}_{e=1}^E$, and assign pseudo-source and pseudo-target according to the similarity between the downstream data and each subset.
For \textbf{domain generalization}, we simply use the first half of the source classes as the pseudo-source and the second half as the pseudo-target.

Within each fold, we lightly fine-tune the baseline model on the pseudo-base/source (training details follow the original paper), then perform a grid search on the pseudo-target to select \(\tilde{\tau}\) and \(\delta\). We flip predictions according to \cref{eq:prior-dominated-region-fixed} and the decision rule, aiming to maximize flip coverage while minimizing the erroneous-flip rate. See the appendix for more details.



\begin{figure}[t]
  \centering
   \includegraphics[width=0.95\linewidth]{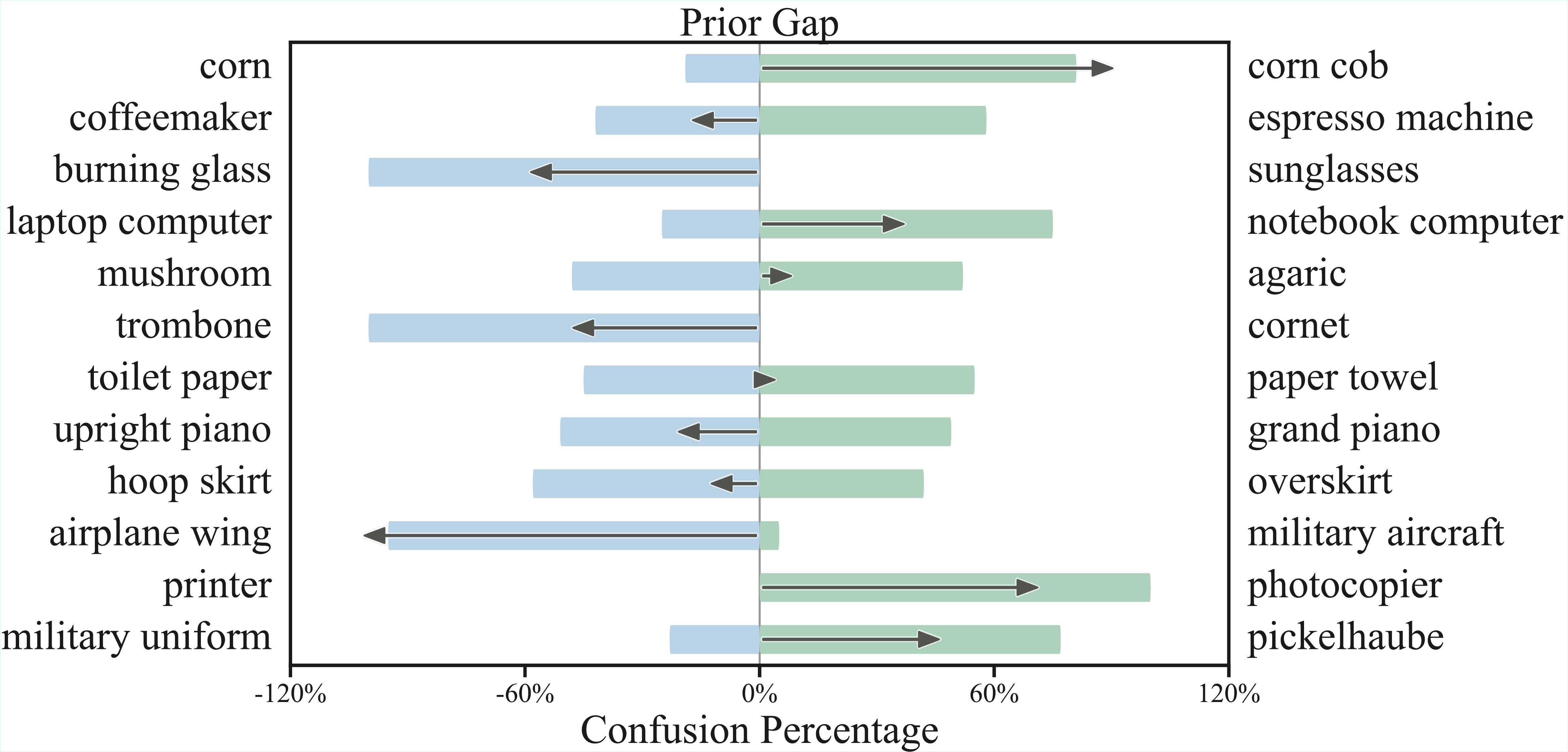}
   \caption{Bars to the right of the y-axis show the fraction of samples from the left class misclassified as the right class, and bars to the left show the converse; arrow length denotes the magnitude of the prior gap.}
   \label{fig:abconsufion}
   \vspace{-0.4cm}
\end{figure}

\subsection{Experimental Results}

\paragraph{Base-to-Novel Generalization}
As shown in \cref{tab:base_to_novel}, we provide detailed results for Base-to-Novel setting across 11 datasets, along with the \emph{balanced harmonic mean (HM)} of their accuracies. Results are encouraging: across four stylistically distinct SOTA baselines, incorporating NeRP consistently improves performance on novel classes. On EuroSAT, several methods gain over 5\%; on Flowers102, DTD, and ImageNet, we observe 0.5–2\% improvements. Because our method adjusts predictions only at inference time, it enhances novel-class accuracy without degrading the performance on base classes.

\vspace{-0.3cm}
\paragraph{Cross-Dataset Evaluation}
\cref{tab:cross_dataset} shows our NeRP method performs well in the case of cross-dataset generalization. Across the four compared methods, adding NeRP yields an average improvement of ~1\% over 10 datasets, with negligible additional overhead.

\vspace{-0.3cm}
\paragraph{Domain Generalization} 
As summarized in \cref{tab:domain_dataset}, NeRP stands out in domain generalization settings: on nearly every dataset, plugging in NeRP yields immediate gains. This provides compelling evidence that pairing NeRP with existing methods can serve as a ‘free lunch’ for efficient transfer learning.

\subsection{Further Analysis}

\paragraph{Effectiveness of Neutral Priors}
We select the top-ranked misclassified class pairs on ImageNet and evaluate the accuracy of neutral-prior probing under symmetric vs. asymmetric confusions (see \cref{fig:abconsufion}). The prior direction aligns with the model’s bias direction, and its magnitude is small in the symmetric-confusion regime. Across datasets, the top-20\% erroneously predicted asymmetric class-confusion pairs (proportion gap \textgreater{}15\%) align with the prior direction with 79.7\% average consistency.

\begin{figure}[!t]
  \centering
   \includegraphics[width=0.95\linewidth]{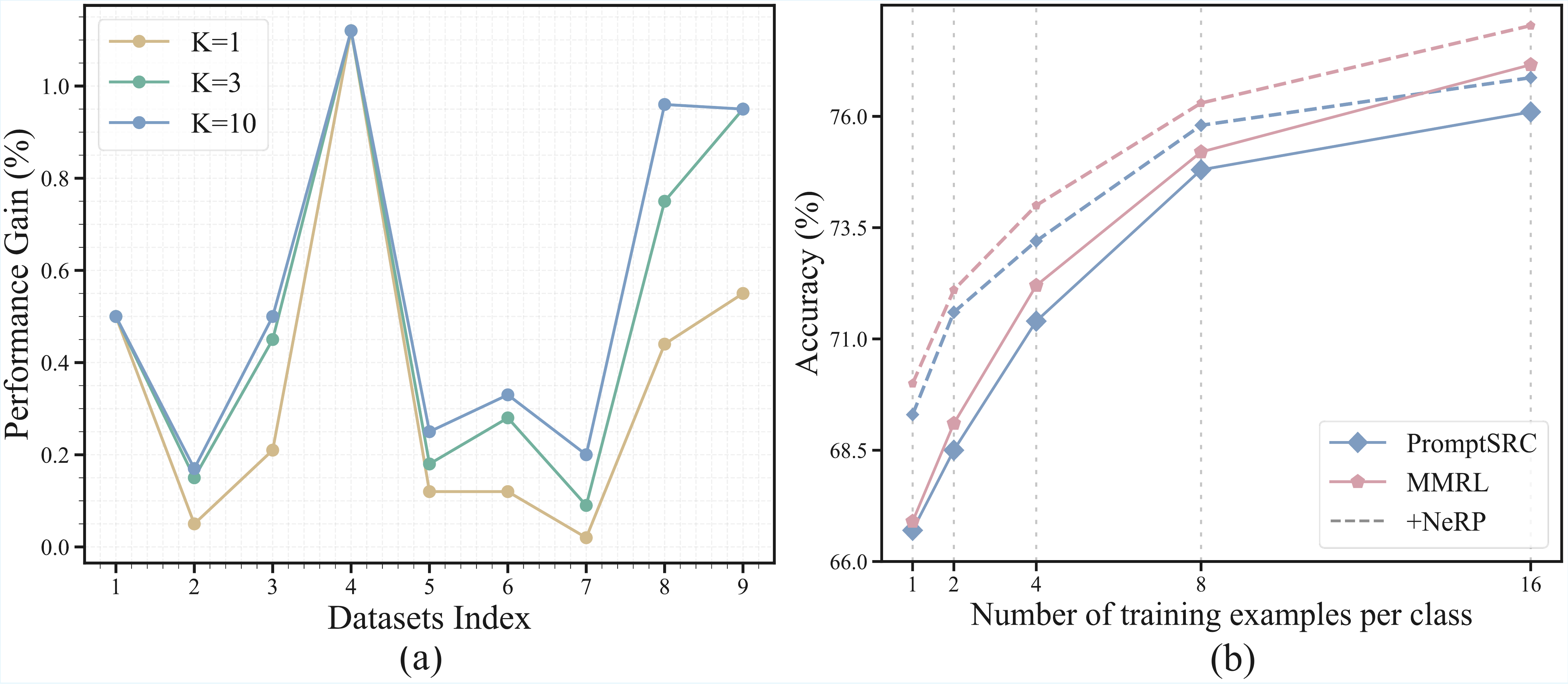}
   \caption{(a) Performance of different $K$. (b) Few-shot performance on unseen classes.}
   \label{fig:few_shot_k}
   \vspace{-0.4cm}
\end{figure}

\vspace{-0.3cm}
\paragraph{Different Calibration/Generation Strategy}
We adopt a unified calibration hyperparameter search as \cref{tab:granularity_folds} shows (10 datasets averaged results). The grid range is set by summarizing model outputs on a small validation subset. Only a single forward pass on \(E\) validation splits is needed; after recording statistics, the search adds negligible overhead. We also study the performance gain of different numbers of confusable pairs across 10 datasets in \cref{fig:few_shot_k} (a) (EuroSAT excluded due to its few classes). On datasets with high inter-class separability, a smaller \(K\) is sufficient; for some datasets, setting \(K=10\) yields larger gains.



\vspace{-0.3cm}
\paragraph{Few-shot Performance}
NeRP remains effective after fine-tuning with fewer samples; results on unseen classes are shown in \cref{fig:few_shot_k} (b). We still have 16 samples per class, but only a few of them are used for fine-tuning, which serve as a validation set to calibrate \(\tilde{\tau}\) and \(\delta\).


%


\begin{table}[!t]
    \centering
    \setlength{\abovecaptionskip}{0.15cm}
    \caption{Ablation on search granularity and number of folds $E$. Metrics are accuracy and flip error rate (FER) (\%).}
    \label{tab:granularity_folds}
    \small
    \setlength{\tabcolsep}{3.5pt}
    \renewcommand{\arraystretch}{0.99}
    \resizebox{\columnwidth}{!}{
        \begin{tabular}{lcc|*{4}{cc}}
        \toprule
        \multicolumn{3}{c}{MMRL+NeRP} & \multicolumn{2}{c}{$E{=}2$} & \multicolumn{2}{c}{$E{=}3$} & \multicolumn{2}{c}{$E{=}4$} & \multicolumn{2}{c}{$E{=}5$} \\
        \cmidrule(lr){4-5}\cmidrule(lr){6-7}\cmidrule(lr){8-9}\cmidrule(lr){10-11}
        Step of & \(\tilde{\tau}\) &  \(\delta\) & Acc. & FER & Acc. & FER & Acc. & FER & Acc. & FER \\
        \midrule
        Coarse  & 1e-2  & 1e-2  & 76.80 & 40.25 & 76.90 & 33.58 & 77.03 & 28.66 & 77.13 & 25.44 \\
        Fine    & 1e-3 & 1e-3 & 76.88 & 36.15 & 77.04 & 28.02 & 77.16 & 23.87 & \textbf{77.32} & \textbf{19.91} \\
        \bottomrule
        \end{tabular}
    }
    \vspace{-0.2cm}
\end{table}

\section{Conclusion}
We identify an asymmetric confusion phenomenon in VLMs under transfer learning and analyze its downstream impact on BNT. Through theoretical and empirical analyses, we further show that a neutral prior can serve as an anchoring signal that reveals model bias. Building on this insight, we introduce NeRP, a plug-and-play correction strategy to mitigate BNT. Extensive experiments demonstrate that NeRP yields consistent improvements across diverse baselines and datasets.


\section*{Impact Statement}

This paper presents work whose goal is to advance the field of Machine
Learning. There are many potential societal consequences of our work, none
which we feel must be specifically highlighted here.


\bibliography{example_paper}

@String(ICLR = {Int. Conf. Learn. Represent.})

@String(AAAI = {AAAI})

@String(ICLR  = {ICLR})

@inproceedings{clip,
  title={Learning transferable visual models from natural language supervision},
  author={Radford, Alec and Kim, Jong Wook and Hallacy, Chris and Ramesh, Aditya and Goh, Gabriel and Agarwal, Sandhini and Sastry, Girish and Askell, Amanda and Mishkin, Pamela and Clark, Jack and others},
  booktitle={International conference on machine learning},
  pages={8748--8763},
  year={2021},
  organization={PMLR}
}

@inproceedings{align,
  title={Scaling up visual and vision-language representation learning with noisy text supervision},
  author={Jia, Chao and Yang, Yinfei and Xia, Ye and Chen, Yi-Ting and Parekh, Zarana and Pham, Hieu and Le, Quoc and Sung, Yun-Hsuan and Li, Zhen and Duerig, Tom},
  booktitle={International conference on machine learning},
  pages={4904--4916},
  year={2021},
  organization={PMLR}
}

@article{filip,
  title={Filip: Fine-grained interactive language-image pre-training},
  author={Yao, Lewei and Huang, Runhui and Hou, Lu and Lu, Guansong and Niu, Minzhe and Xu, Hang and Liang, Xiaodan and Li, Zhenguo and Jiang, Xin and Xu, Chunjing},
  journal={arXiv preprint arXiv:2111.07783},
  year={2021}
}

@article{flamingo,
  title={Flamingo: a visual language model for few-shot learning},
  author={Alayrac, Jean-Baptiste and Donahue, Jeff and Luc, Pauline and Miech, Antoine and Barr, Iain and Hasson, Yana and Lenc, Karel and Mensch, Arthur and Millican, Katherine and Reynolds, Malcolm and others},
  journal={Advances in neural information processing systems},
  volume={35},
  pages={23716--23736},
  year={2022}
}

@inproceedings{vila,
  title={Vila: On pre-training for visual language models},
  author={Lin, Ji and Yin, Hongxu and Ping, Wei and Molchanov, Pavlo and Shoeybi, Mohammad and Han, Song},
  booktitle={Proceedings of the IEEE/CVF Conference on Computer Vision and Pattern Recognition},
  pages={26689--26699},
  year={2024}
}

@article{coop,
  title={Learning to prompt for vision-language models},
  author={Zhou, Kaiyang and Yang, Jingkang and Loy, Chen Change and Liu, Ziwei},
  journal={International Journal of Computer Vision},
  volume={130},
  number={9},
  pages={2337--2348},
  year={2022},
  publisher={Springer}
}

@inproceedings{cocoop,
  title={Conditional prompt learning for vision-language models},
  author={Zhou, Kaiyang and Yang, Jingkang and Loy, Chen Change and Liu, Ziwei},
  booktitle={Proceedings of the IEEE/CVF conference on computer vision and pattern recognition},
  pages={16816--16825},
  year={2022}
}

@inproceedings{proda,
  title={Prompt distribution learning},
  author={Lu, Yuning and Liu, Jianzhuang and Zhang, Yonggang and Liu, Yajing and Tian, Xinmei},
  booktitle={Proceedings of the IEEE/CVF Conference on Computer Vision and Pattern Recognition},
  pages={5206--5215},
  year={2022}
}

@inproceedings{maple,
  title={Maple: Multi-modal prompt learning},
  author={Khattak, Muhammad Uzair and Rasheed, Hanoona and Maaz, Muhammad and Khan, Salman and Khan, Fahad Shahbaz},
  booktitle={Proceedings of the IEEE/CVF Conference on Computer Vision and Pattern Recognition},
  pages={19113--19122},
  year={2023}
}

@inproceedings{prograd,
  title={Prompt-aligned gradient for prompt tuning},
  author={Zhu, Beier and Niu, Yulei and Han, Yucheng and Wu, Yue and Zhang, Hanwang},
  booktitle={Proceedings of the IEEE/CVF International Conference on Computer Vision},
  pages={15659--15669},
  year={2023}
}

@inproceedings{kgcoop,
  title={Visual-language prompt tuning with knowledge-guided context optimization},
  author={Yao, Hantao and Zhang, Rui and Xu, Changsheng},
  booktitle={Proceedings of the IEEE/CVF conference on computer vision and pattern recognition},
  pages={6757--6767},
  year={2023}
}

@inproceedings{promptsrc,
  title={Self-regulating prompts: Foundational model adaptation without forgetting},
  author={Khattak, Muhammad Uzair and Wasim, Syed Talal and Naseer, Muzammal and Khan, Salman and Yang, Ming-Hsuan and Khan, Fahad Shahbaz},
  booktitle={Proceedings of the IEEE/CVF International Conference on Computer Vision},
  pages={15190--15200},
  year={2023}
}

@inproceedings{tcp,
  title={TCP: Textual-based Class-aware Prompt tuning for Visual-Language Model},
  author={Yao, Hantao and Zhang, Rui and Xu, Changsheng},
  booktitle={Proceedings of the IEEE/CVF Conference on Computer Vision and Pattern Recognition},
  pages={23438--23448},
  year={2024}
}

@article{provp,
  title={Progressive visual prompt learning with contrastive feature re-formation},
  author={Xu, Chen and Zhu, Yuhan and Shen, Haocheng and Chen, Boheng and Liao, Yixuan and Chen, Xiaoxin and Wang, Limin},
  journal={International Journal of Computer Vision},
  pages={1--16},
  year={2024},
  publisher={Springer}
}

@article{clip-adapter,
  title={Clip-adapter: Better vision-language models with feature adapters},
  author={Gao, Peng and Geng, Shijie and Zhang, Renrui and Ma, Teli and Fang, Rongyao and Zhang, Yongfeng and Li, Hongsheng and Qiao, Yu},
  journal={International Journal of Computer Vision},
  volume={132},
  number={2},
  pages={581--595},
  year={2024},
  publisher={Springer}
}

@inproceedings{mma,
  title={MMA: Multi-Modal Adapter for Vision-Language Models},
  author={Yang, Lingxiao and Zhang, Ru-Yuan and Wang, Yanchen and Xie, Xiaohua},
  booktitle={Proceedings of the IEEE/CVF Conference on Computer Vision and Pattern Recognition},
  pages={23826--23837},
  year={2024}
}

@inproceedings{coprompt,
  title={Consistency-guided Prompt Learning for Vision-Language Models},
  author={Roy, Shuvendu and Etemad, Ali},
  booktitle=ICLR,
  year={2024}
}

@inproceedings{promptkd,
  title={Promptkd: Unsupervised prompt distillation for vision-language models},
  author={Li, Zheng and Li, Xiang and Fu, Xinyi and Zhang, Xin and Wang, Weiqiang and Chen, Shuo and Yang, Jian},
  booktitle={Proceedings of the IEEE/CVF Conference on Computer Vision and Pattern Recognition},
  pages={26617--26626},
  year={2024}
}

@inproceedings{imagenet,
  title={Imagenet: A large-scale hierarchical image database},
  author={Deng, Jia and Dong, Wei and Socher, Richard and Li, Li-Jia and Li, Kai and Fei-Fei, Li},
  booktitle={2009 IEEE conference on computer vision and pattern recognition},
  pages={248--255},
  year={2009},
  organization={Ieee}
}

@inproceedings{sun397,
  title={Sun database: Large-scale scene recognition from abbey to zoo},
  author={Xiao, Jianxiong and Hays, James and Ehinger, Krista A and Oliva, Aude and Torralba, Antonio},
  booktitle={2010 IEEE computer society conference on computer vision and pattern recognition},
  pages={3485--3492},
  year={2010},
  organization={IEEE}
}

@inproceedings{caltech101,
  title={Learning generative visual models from few training examples: An incremental bayesian approach tested on 101 object categories},
  author={Fei-Fei, Li and Fergus, Rob and Perona, Pietro},
  booktitle={2004 conference on computer vision and pattern recognition workshop},
  pages={178--178},
  year={2004},
  organization={IEEE}
}

@inproceedings{oxford_pets,
  title={Cats and dogs},
  author={Parkhi, Omkar M and Vedaldi, Andrea and Zisserman, Andrew and Jawahar, CV},
  booktitle={2012 IEEE conference on computer vision and pattern recognition},
  pages={3498--3505},
  year={2012},
  organization={IEEE}
}

@inproceedings{stanford_cars,
  title={3d object representations for fine-grained categorization},
  author={Krause, Jonathan and Stark, Michael and Deng, Jia and Fei-Fei, Li},
  booktitle={Proceedings of the IEEE international conference on computer vision workshops},
  pages={554--561},
  year={2013}
}

@inproceedings{flowers102,
  title={Automated flower classification over a large number of classes},
  author={Nilsback, Maria-Elena and Zisserman, Andrew},
  booktitle={2008 Sixth Indian conference on computer vision, graphics \& image processing},
  pages={722--729},
  year={2008},
  organization={IEEE}
}

@article{ucf101,
  title={UCF101: A Dataset of 101 Human Actions Classes From Videos in The Wild},
  author={Khurram Soomro and Amir Zamir and Mubarak Shah},
  journal={ArXiv},
  year={2012},
  volume={abs/1212.0402}
}

@inproceedings{dtd,
  title={Describing textures in the wild},
  author={Cimpoi, Mircea and Maji, Subhransu and Kokkinos, Iasonas and Mohamed, Sammy and Vedaldi, Andrea},
  booktitle={Proceedings of the IEEE conference on computer vision and pattern recognition},
  pages={3606--3613},
  year={2014}
}

@article{eurosat,
  title={Eurosat: A novel dataset and deep learning benchmark for land use and land cover classification},
  author={Helber, Patrick and Bischke, Benjamin and Dengel, Andreas and Borth, Damian},
  journal={IEEE Journal of Selected Topics in Applied Earth Observations and Remote Sensing},
  volume={12},
  number={7},
  pages={2217--2226},
  year={2019},
  publisher={IEEE}
}

@inproceedings{food101,
  title={Food-101--mining discriminative components with random forests},
  author={Bossard, Lukas and Guillaumin, Matthieu and Van Gool, Luc},
  booktitle={Computer vision--ECCV 2014: 13th European conference, zurich, Switzerland, September 6-12, 2014, proceedings, part VI 13},
  pages={446--461},
  year={2014},
  organization={Springer}
}

@article{fgvc_aircraft,
  title={Fine-grained visual classification of aircraft},
  author={Maji, Subhransu and Rahtu, Esa and Kannala, Juho and Blaschko, Matthew and Vedaldi, Andrea},
  journal={arXiv preprint arXiv:1306.5151},
  year={2013}
}

@inproceedings{imagenetv2,
  title={Do imagenet classifiers generalize to imagenet?},
  author={Recht, Benjamin and Roelofs, Rebecca and Schmidt, Ludwig and Shankar, Vaishaal},
  booktitle={International conference on machine learning},
  pages={5389--5400},
  year={2019},
  organization={PMLR}
}

@article{imagenet_sketch,
  title={Learning robust global representations by penalizing local predictive power},
  author={Wang, Haohan and Ge, Songwei and Lipton, Zachary and Xing, Eric P},
  journal={Advances in Neural Information Processing Systems},
  volume={32},
  year={2019}
}

@inproceedings{imagenet_a,
  title={Natural adversarial examples},
  author={Hendrycks, Dan and Zhao, Kevin and Basart, Steven and Steinhardt, Jacob and Song, Dawn},
  booktitle={Proceedings of the IEEE/CVF conference on computer vision and pattern recognition},
  pages={15262--15271},
  year={2021}
}

@inproceedings{imagenet_r,
  title={The many faces of robustness: A critical analysis of out-of-distribution generalization},
  author={Hendrycks, Dan and Basart, Steven and Mu, Norman and Kadavath, Saurav and Wang, Frank and Dorundo, Evan and Desai, Rahul and Zhu, Tyler and Parajuli, Samyak and Guo, Mike and others},
  booktitle={Proceedings of the IEEE/CVF international conference on computer vision},
  pages={8340--8349},
  year={2021}
}

@inproceedings{dept,
  title={Dept: Decoupled prompt tuning},
  author={Zhang, Ji and Wu, Shihan and Gao, Lianli and Shen, Heng Tao and Song, Jingkuan},
  booktitle={Proceedings of the IEEE/CVF Conference on Computer Vision and Pattern Recognition},
  pages={12924--12933},
  year={2024}
}

@inproceedings{dpc,
  title={Dpc: Dual-prompt collaboration for tuning vision-language models},
  author={Li, Haoyang and Wang, Liang and Wang, Chao and Jiang, Jing and Peng, Yan and Long, Guodong},
  booktitle={Proceedings of the Computer Vision and Pattern Recognition Conference},
  pages={25623--25632},
  year={2025}
}

@inproceedings{mmrl,
  title={Mmrl: Multi-modal representation learning for vision-language models},
  author={Guo, Yuncheng and Gu, Xiaodong},
  booktitle={Proceedings of the Computer Vision and Pattern Recognition Conference},
  pages={25015--25025},
  year={2025}
}

@inproceedings{skipt,
  title={Skip tuning: Pre-trained vision-language models are effective and efficient adapters themselves},
  author={Wu, Shihan and Zhang, Ji and Zeng, Pengpeng and Gao, Lianli and Song, Jingkuan and Shen, Heng Tao},
  booktitle={Proceedings of the Computer Vision and Pattern Recognition Conference},
  pages={14723--14732},
  year={2025}
}

@inproceedings{nlprompt,
  title={NLPrompt: Noise-Label Prompt Learning for Vision-Language Models},
  author={Pan, Bikang and Li, Qun and Tang, Xiaoying and Huang, Wei and Fang, Zhen and Liu, Feng and Wang, Jingya and Yu, Jingyi and Shi, Ye},
  booktitle={Proceedings of the Computer Vision and Pattern Recognition Conference},
  pages={19963--19973},
  year={2025}
}

@inproceedings{surpl,
  title={Surrogate Prompt Learning: Towards Efficient and Diverse Prompt Learning for Vision-Language Models},
  author={Liu, Liangchen and Wang, Nannan and Yang, Xi and Gao, Xinbo and Liu, Tongliang},
  booktitle={Forty-second International Conference on Machine Learning},
  year={2025}
}

@inproceedings{caspl,
  title={Cascade prompt learning for vision-language model adaptation},
  author={Wu, Ge and Zhang, Xin and Li, Zheng and Chen, Zhaowei and Liang, Jiajun and Yang, Jian and Li, Xiang},
  booktitle={European Conference on Computer Vision},
  pages={304--321},
  year={2024},
  organization={Springer}
}

@inproceedings{sobit,
  title={Identifying implicit social biases in vision-language models},
  author={Hamidieh, Kimia and Zhang, Haoran and Gerych, Walter and Hartvigsen, Thomas and Ghassemi, Marzyeh},
  booktitle={Proceedings of the AAAI/ACM Conference on AI, Ethics, and Society},
  volume={7},
  pages={547--561},
  year={2024}
}

@article{clipthebias,
  title={Clip the bias: How useful is balancing data in multimodal learning?},
  author={Alabdulmohsin, Ibrahim and Wang, Xiao and Steiner, Andreas and Goyal, Priya and D'Amour, Alexander and Zhai, Xiaohua},
  journal={arXiv preprint arXiv:2403.04547},
  year={2024}
}

@article{sizenoteq,
  title={A Comprehensive Social Bias Audit of Contrastive Vision Language Models},
  author={Sahili, Zahraa Al and Patras, Ioannis and Purver, Matthew},
  journal={arXiv preprint arXiv:2501.13223},
  year={2025}
}

@article{intrinsic,
  title={Intrinsic bias is predicted by pretraining data and correlates with downstream performance in vision-language encoders},
  author={Ghate, Kshitish and Slaughter, Isaac and Wilson, Kyra and Diab, Mona and Caliskan, Aylin},
  journal={arXiv preprint arXiv:2502.07957},
  year={2025}
}

@article{sober,
  title={A sober look at the robustness of clips to spurious features},
  author={Wang, Qizhou and Lin, Yong and Chen, Yongqiang and Schmidt, Ludwig and Han, Bo and Zhang, Tong},
  journal={Advances in Neural Information Processing Systems},
  volume={37},
  pages={122484--122523},
  year={2024}
}

@article{closer,
  title={A closer look at the robustness of contrastive language-image pre-training (clip)},
  author={Tu, Weijie and Deng, Weijian and Gedeon, Tom},
  journal={Advances in Neural Information Processing Systems},
  volume={36},
  pages={13678--13691},
  year={2023}
}

@inproceedings{bgprompt,
  title={Learning background prompts to discover implicit knowledge for open vocabulary object detection},
  author={Li, Jiaming and Zhang, Jiacheng and Li, Jichang and Li, Ge and Liu, Si and Lin, Liang and Li, Guanbin},
  booktitle={Proceedings of the IEEE/CVF Conference on Computer Vision and Pattern Recognition},
  pages={16678--16687},
  year={2024}
}

@article{rsclip,
  title={RS-CLIP: Zero shot remote sensing scene classification via contrastive vision-language supervision},
  author={Li, Xiang and Wen, Congcong and Hu, Yuan and Zhou, Nan},
  journal={International Journal of Applied Earth Observation and Geoinformation},
  volume={124},
  pages={103497},
  year={2023},
  publisher={Elsevier}
}

@article{transfer2,
  title={Supervision exists everywhere: A data efficient contrastive language-image pre-training paradigm},
  author={Li, Yangguang and Liang, Feng and Zhao, Lichen and Cui, Yufeng and Ouyang, Wanli and Shao, Jing and Yu, Fengwei and Yan, Junjie},
  journal={arXiv preprint arXiv:2110.05208},
  year={2021}
}

@inproceedings{transfer3,
  title={Flava: A foundational language and vision alignment model},
  author={Singh, Amanpreet and Hu, Ronghang and Goswami, Vedanuj and Couairon, Guillaume and Galuba, Wojciech and Rohrbach, Marcus and Kiela, Douwe},
  booktitle={Proceedings of the IEEE/CVF conference on computer vision and pattern recognition},
  pages={15638--15650},
  year={2022}
}

@article{qwen2,
  title={Qwen2.5-1m technical report},
  author={Yang, An and Yu, Bowen and Li, Chengyuan and Liu, Dayiheng and Huang, Fei and Huang, Haoyan and Jiang, Jiandong and Tu, Jianhong and Zhang, Jianwei and Zhou, Jingren and others},
  journal={arXiv preprint arXiv:2501.15383},
  year={2025}
}

@article{lora1,
  title={Lora: Low-rank adaptation of large language models.},
  author={Hu, Edward J and Shen, Yelong and Wallis, Phillip and Allen-Zhu, Zeyuan and Li, Yuanzhi and Wang, Shean and Wang, Lu and Chen, Weizhu and others},
  journal={ICLR},
  volume={1},
  number={2},
  pages={3},
  year={2022}
}

@article{lora2,
  title={Towards a unified view of parameter-efficient transfer learning},
  author={He, Junxian and Zhou, Chunting and Ma, Xuezhe and Berg-Kirkpatrick, Taylor and Neubig, Graham},
  journal={arXiv preprint arXiv:2110.04366},
  year={2021}
}

@article{lora3,
  title={Universality and limitations of prompt tuning},
  author={Wang, Yihan and Chauhan, Jatin and Wang, Wei and Hsieh, Cho-Jui},
  journal={Advances in Neural Information Processing Systems},
  volume={36},
  pages={75623--75643},
  year={2023}
}

@inproceedings{vmf1,
  title={Improving zero-shot generalization for clip with variational adapter},
  author={Lu, Ziqian and Shen, Fengli and Liu, Mushui and Yu, Yunlong and Li, Xi},
  booktitle={European Conference on Computer Vision},
  pages={328--344},
  year={2024},
  organization={Springer}
}

@article{vmf2,
  title={Clustering on the Unit Hypersphere using von Mises-Fisher Distributions.},
  author={Banerjee, Arindam and Dhillon, Inderjit S and Ghosh, Joydeep and Sra, Suvrit and Ridgeway, Greg},
  journal={Journal of Machine Learning Research},
  volume={6},
  number={9},
  year={2005}
}

@article{tian2026sampling,
  title={Sampling Control for Imbalanced Calibration in Semi-Supervised Learning},
  author={Tian, Senmao and Wei, Xiang and Zhang, Shunli},
  journal={Proceedings of the AAAI Conference on Artificial Intelligence},
  volume={40},
  number={31},
  pages={25914--25922},
  year={2026}
}

@article{a1,
  title={MMErroR: A Benchmark for Erroneous Reasoning in Vision-Language Models},
  author={Shi, Yang and Xie, Yifeng and Guo, Minzhe and Lu, Liangsi and Huang, Mingxuan and Wang, Jingchao and Zhu, Zhihong and Xu, Boyan and Huang, Zhiqi},
  journal={arXiv preprint arXiv:2601.03331},
  year={2026}
}

@article{a2,
  title={ChordEdit: One-Step Low-Energy Transport for Image Editing},
  author={Lu, Liangsi and Chen, Xuhang and Guo, Minzhe and Li, Shichu and Wang, Jingchao and Shi, Yang},
  journal={arXiv preprint arXiv:2602.19083},
  year={2026}
}

@inproceedings{
a3,
title={M{\texttwosuperior}{IV}: Towards Efficient and Fine-grained Multimodal In-Context Learning via Representation Engineering},
author={Yanshu Li and Yi Cao and Hongyang He and Qisen Cheng and Xiang Fu and Xi Xiao and Tianyang Wang and Ruixiang Tang},
booktitle={Second Conference on Language Modeling},
year={2025},
}

@article{a4,
  title={Make LVLMs Focus: Context-Aware Attention Modulation for Better Multimodal In-Context Learning},
  author={Li, Yanshu and Yang, Jianjiang and Yang, Ziteng and Li, Bozheng and Han, Ligong and He, Hongyang and Yao, Zhengtao and Chen, Yingjie Victor and Fei, Songlin and Liu, Dongfang and others},
  journal={Proceedings of the AAAI Conference on Artificial Intelligence},
  volume={40},
  number={8},
  pages={6610--6618},
  year={2026}
}

@article{a5,
  title={Large Vision-Language Models Get Lost in Attention},
  author={Xi, Gongli and Tian, Ye and Yang, Mengyu and Yi, Huahui and Lin, Liang and Hao, Xiaoshuai and Wang, Kun and Wang, Wendong},
  journal={arXiv preprint arXiv:2605.05668},
  year={2026}
}

@article{a6,
  title={Gated Relational Alignment via Confidence-based Distillation for Efficient VLMs},
  author={Chen, Yanlong and Habibian, Amirhossein and Benini, Luca and Li, Yawei},
  journal={arXiv preprint arXiv:2601.22709},
  year={2026}
}
\bibliographystyle{icml2026}



\end{document}